\documentclass[letterpaper,journal]{IEEEtran}
\usepackage{amsmath,amsfonts}
\usepackage{algorithmic}
\usepackage{algorithm}
\usepackage{array}
\usepackage[caption=false,font=normalsize,labelfont=sf,textfont=sf]{subfig}
\usepackage{textcomp}
\usepackage{stfloats}
\usepackage{url}
\usepackage{verbatim}
\usepackage{graphicx}
\usepackage{cite}
\usepackage{orcidlink}
\usepackage{enumitem}
\usepackage{booktabs}
\usepackage{tabularx}
\usepackage{xcolor}
\usepackage[normalem]{ulem}
\hypersetup{hidelinks}
\newif\ifshowchanges
\ifdefined\reviewversion
  \showchangestrue
\else
  \showchangesfalse
\fi
\ifshowchanges
  
  \long\def\rev#1{\textcolor{red}{#1}}
\else
  
  \long\def\rev#1{\textcolor{black}{#1}}
\fi
\newcommand{\del}[1]{}
\newcommand{\changedcolor}{}
\newcommand{\reviewfloatcolor}{\ifshowchanges\color{red}\else\color{black}\fi}
\begin{document}
\bstctlcite{IEEEauthorlistControl}

\title{Large Language Models in Transportation Systems Management and Operations: From Text Reasoning to Multi-modal Decision Support}

\author{{%
    Siyan Li$^{*\orcidlink{0009-0000-2222-963X}}$,~\IEEEmembership{Student~Member,~IEEE},
    Zehao Wang$^{\orcidlink{0009-0006-8797-8571}}$,~\IEEEmembership{Student~Member,~IEEE},
    Jiachen Li$^{\orcidlink{0000-0002-4883-697X}}$,~\IEEEmembership{Member,~IEEE},
    Kanok Boriboonsomsin$^{\orcidlink{0000-0003-2558-5343}}$,~\IEEEmembership{Member,~IEEE},
    Matthew J. Barth$^{\orcidlink{0000-0002-4735-5859}}$,~\IEEEmembership{Fellow,~IEEE},
    Guoyuan Wu$^{\orcidlink{0000-0001-6707-6366}}$,~\IEEEmembership{Senior~Member,~IEEE}%
}

\thanks{Siyan Li, Kanok Boriboonsomsin, Matthew J. Barth, and Guoyuan Wu are with the Bourns College of Engineering, Center for Environmental Research and Technology, University of California at Riverside, CA, USA.}

\thanks{Zehao Wang is with the Bourns College of Engineering, Computer Science and Engineering Department, University of California at Riverside, CA, USA.}

\thanks{Jiachen Li is with the Bourns College of Engineering, Electrical and Computer Engineering Department, University of California at Riverside, CA, USA.}

\thanks{$^{*}$Corresponding author: Siyan Li; Email: sli442@ucr.edu}

}


\maketitle

\begin{abstract}
Transportation systems management and operations (TSMO) increasingly depends on timely interpretation of heterogeneous data, from various sensor streams, incident reports, traveler feedback, and visual observations. Large language models (LLMs), including emerging multi-modal large language models (MM-LLMs), provide a mechanism for integrating these structured and unstructured inputs into operator-facing decision support. Informed by the preferred reporting items for systematic reviews and meta-analyses (PRISMA), this survey reviews LLM- and MM-LLM-based applications in TSMO across transportation operations \& services (supply) and mobility \& fleet services (demand), with data, modeling \& decision support as a cross-cutting supporting layer. \rev{Through source-specific comparisons, we synthesize current studies while distinguishing operationally oriented applications from prototypes and emerging concepts, and separating validation maturity from action authority.} We further identify recurring challenges in data heterogeneity, real-time inference, explainability, multi-modal fusion, and governance. Finally, we outline existing gaps and future directions in localized adaptation, edge deployment, benchmarking, and cross-agency collaboration. Overall, LLM-based systems offer a decision-support layer, \rev{with the value of integrating heterogeneous text, visual, and sensor inputs depending on task-specific validation and operational safeguards.}
\end{abstract}


\section{Introduction}
\label{intro}
\subsection{Background}
Transportation systems management and operations (TSMO) is dedicated to optimizing the performance of existing transportation networks by coordinating infrastructure, vehicles, road users, and information flows through advanced operational strategies \cite{FederalHighwayAdministration2025TransportationOperations}. Real-time decision-support platforms respond to dynamic traffic conditions, stakeholder demands, and multi-modal mobility patterns \cite{Khalil2024AdvancedChallenges}.

\begin{figure*}[!t]\reviewfloatcolor
\centering
\includegraphics[width=\linewidth]{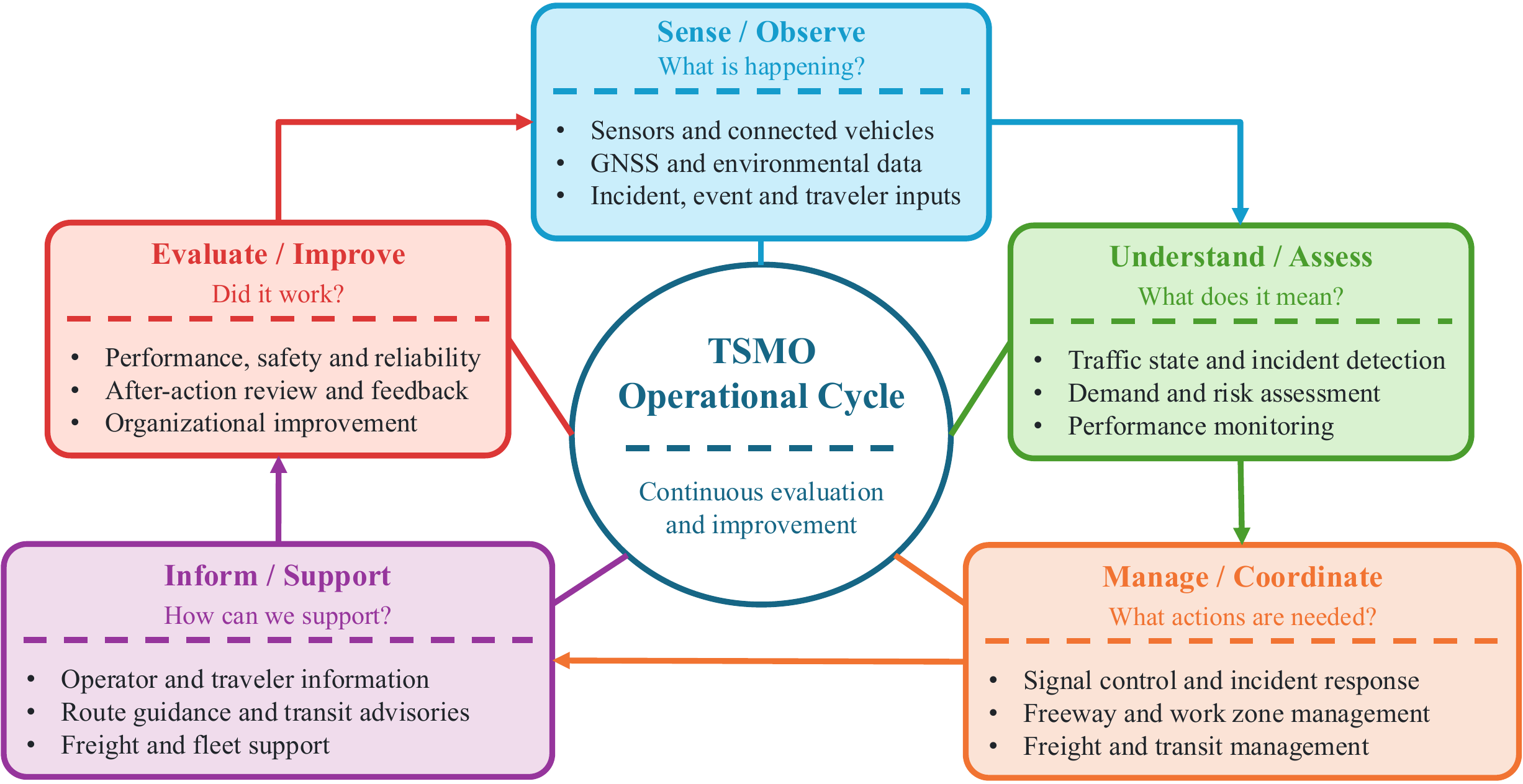}
\caption{\rev{TSMO operational cycle linking observation, assessment, coordination and information support with feedback for continuous, governed improvement. GNSS denotes global navigation satellite system.}}
\label{Fig:TSMO_Cycle}
\end{figure*}

The modern transportation environment produces a rich mix of observations, from sensors and probe-vehicle telemetry to incident narratives, crowd-sourced alerts, and live imagery \cite{Sarwatt2024MetaverseDirections}. Textual and visual reports convey localized hazards, weather impacts and crowd behavior that can complement numerical and rule-based analyses. Figure \ref{Fig:TSMO_Cycle} illustrates the general operational cycle of TSMO.

Recent advances in artificial intelligence (AI) and large language models (LLMs) \cite{Brown2020LanguageLearners}, especially multi-modal large language models (MM-LLMs), extend beyond text generation to joint processing of language, vision, and sensor inputs \cite{Yin2023AModels}. Connecting heterogeneous modalities to a language backbone offers mechanisms for detecting anomalies, anticipating disruptions and generating context-aware recommendations. Benefits over existing TSMO methods require task-specific evaluation.

Incomplete integration of these sources can constrain operational decision-making. This survey focuses specifically on system-level TSMO applications of LLMs and synthesizes how they are being positioned within transportation operations \& services (supply), mobility \& fleet services (demand), and data, modeling \& decision support workflows.

\subsection{Study Scope}
\label{sec:protocol}
\rev{This review uses the preferred reporting items for systematic reviews and meta-analyses (PRISMA) 2020 statement to guide the reporting of literature identification and selection \cite{Page2021TheReviews}.} The scope emphasizes system-level transportation operations, mobility services and their supporting analytical workflows. \rev{It combines peer-reviewed papers with relevant preprints, technical reports and agency or practitioner publications.} Earlier foundational methods and conventional baselines provide context.

The search combines existing references and targeted checks with the bibliographic strategy in Table~\ref{tab:search_strategy_audit}; Table~\ref{tab:eligibility_criteria_audit} summarizes eligibility. \rev{Primary applications involve a pretrained generative LLM component or a detailed proposed role in a TSMO workflow. Operational forecasting, incident interpretation, service assistance and simulation can qualify through offline evaluation. Generic travel-behavior models and automated-driving studies provide supporting context when their evaluated tasks fall outside this operational scope.}

\begin{table*}[t]\reviewfloatcolor
\centering
\caption{\rev{Literature search settings and query strings.}}
\label{tab:search_strategy_audit}
\small
\setlength{\tabcolsep}{5pt}
\renewcommand{\arraystretch}{1.10}
\begin{tabularx}{\textwidth}{@{}p{0.18\textwidth}>{\raggedright\arraybackslash}X@{}}
\toprule
\textbf{Setting} & \textbf{Specification} \\
\midrule
Search Source & OpenAlex Works application programming interface (API), indexed titles and abstracts \\
\addlinespace[2.5pt]
Publication Period & January 1, 2023--August 31, 2026 \\
\addlinespace[2.5pt]
Additional Sources & Existing references and targeted checks, including reviewer suggestions \\
\addlinespace[2.5pt]
Filters & English search terms; no language, publication-type or open-access filter \\
\bottomrule
\end{tabularx}
\medskip
\begin{tabularx}{\textwidth}{@{}p{0.18\textwidth}>{\raggedright\arraybackslash}X@{}}
\toprule
\textbf{Query Family} & \textbf{Exact Query String} \\
\midrule
Operations & \texttt{large language model transportation operations} \\
\addlinespace[2.5pt]
Traffic Management & \texttt{large language model traffic management} \\
\addlinespace[2.5pt]
Multi-modal TSMO & \texttt{multi-modal large language model intelligent transportation} \\
\addlinespace[2.5pt]
Public Transit & \texttt{large language model public transit} \\
\addlinespace[2.5pt]
Signal Control & \texttt{large language model traffic signal control} \\
\addlinespace[2.5pt]
Incidents & \texttt{large language model traffic incident management} \\
\addlinespace[2.5pt]
Freight and Routing & \texttt{large language model freight logistics routing} \\
\addlinespace[2.5pt]
Simulation & \texttt{large language model traffic simulation} \\
\addlinespace[2.5pt]
Benchmarking & \texttt{large language model transportation benchmark} \\
\addlinespace[2.5pt]
Generative AI & \texttt{generative AI transportation operations} \\
\addlinespace[2.5pt]
Vision--language & \texttt{vision language model traffic transportation} \\
\addlinespace[2.5pt]
Foundation Models & \texttt{foundation model intelligent transportation systems} \\
\bottomrule
\end{tabularx}
\end{table*}

\begin{table*}[t]\reviewfloatcolor
\centering
\caption{\rev{Literature eligibility and coding criteria.}}
\label{tab:eligibility_criteria_audit}
\small
\setlength{\tabcolsep}{5pt}
\renewcommand{\arraystretch}{1.10}
\begin{tabularx}{\textwidth}{@{}p{0.19\textwidth}>{\raggedright\arraybackslash}X@{}}
\toprule
\textbf{Dimension} & \textbf{Criterion} \\
\midrule
Date and Version & Primary application reports available within the search period, including eligible preprints and online publications. Earlier foundations and baselines provide context. \\
\addlinespace[3pt]
Model Component & A pretrained generative language backbone in an implemented system, or a defined generative LLM function in a conceptual design. Record numerical decoder adaptation separately from conversational and native multi-modal use. Encoder-only and conventional methods provide context. \\
\addlinespace[3pt]
Operational Task & Road-transport or urban-transit operations, service support, operational analysis, simulation or task evaluation. Offline forecasting and incident interpretation are eligible. \\
\addlinespace[3pt]
Demand and Routing & Forecasting, disruption response, routing or dispatch linked to a defined service or network task. Generic mode-choice, recognition and activity-generation models provide context unless evaluated in such a task. \\
\addlinespace[3pt]
Automated Driving & Primarily ego-driving tasks provide transfer evidence. Mixed-scope reports are assessed by their eligible incident-analysis or operational component. \\
\addlinespace[3pt]
Source Type & Journal and conference papers, preprints, technical reports and relevant agency/practitioner publications. Detailed operational designs can qualify as M0; broader visions and reviews provide context. \\
\addlinespace[3pt]
Evidence Access & Primary inclusion requires substantive inspected-body evidence for the claimed task, model role and source identity. Records with insufficient information or access remain unresolved. \\
\addlinespace[3pt]
Evaluation & Record the tested LLM component, dataset or scenario, comparator, principal result, maturity and limitations. Assess publication status, action authority and methodological quality separately. \\
\addlinespace[3pt]
Report Identity & Link revisions, related publications and reused extractions in the review records. Preserve shared-experiment relationships when reporting counts. \\
\bottomrule
\end{tabularx}
\end{table*}

\rev{The synthesis compares each model's function, inputs, operational task, evaluation setting, principal findings and limitations. Validation maturity is coded for the LLM component using M0--M4, as shown in Table~\ref{tab:maturity_report_counts}. An identified reference can be labeled data, expected tool output or a competing method. Publication status and action authority are recorded separately from methodological quality. Qualitative appraisal of the selected comparisons considers matched baselines, data and setup transparency, uncertainty reporting, and constraint verifications, including whether gains can be attributed to the LLM component. }

\begin{table}[!tbp]\reviewfloatcolor
\centering
\caption{\rev{Validation maturity of LLM components across the primary report entries.}}
\label{tab:maturity_report_counts}
\small
\setlength{\tabcolsep}{4pt}
\renewcommand{\arraystretch}{1.10}
\begin{tabularx}{\columnwidth}{@{}l>{\raggedright\arraybackslash}Xr@{}}
\toprule
\textbf{Stage} & \textbf{Evaluation setting} & \textbf{Count} \\
\midrule
M0 & Conceptual design & 3 \\
\addlinespace[3pt]
M1 & Implemented illustration without a quantitative reference test & 26 \\
\addlinespace[3pt]
M2 & Offline or simulation test with a defined quantitative reference & 97 \\
\addlinespace[3pt]
M3* & Bounded live pilot & 1 \\
\addlinespace[3pt]
M4 & Documented sustained routine operation & 0 \\
\midrule
Total & Report entries & 127 \\
\bottomrule
\end{tabularx}
\par\vspace{5pt}\noindent
\begin{minipage}{\columnwidth}
\footnotesize
*The M3 entry describes Traffic-R1's author-reported supervised pilot.
\end{minipage}
\end{table}

\rev{Screening, extraction and source-consistency checks were AI-assisted, with decisions archived for author verification. The protocol developed during the review, with refinements documented in the retained review records. Coverage is limited by the single-index search, English query terms and uneven source access; related reports may also share experimental evidence.}

\begin{figure}[!htbp]\reviewfloatcolor
\centering
\includegraphics[width=\columnwidth]{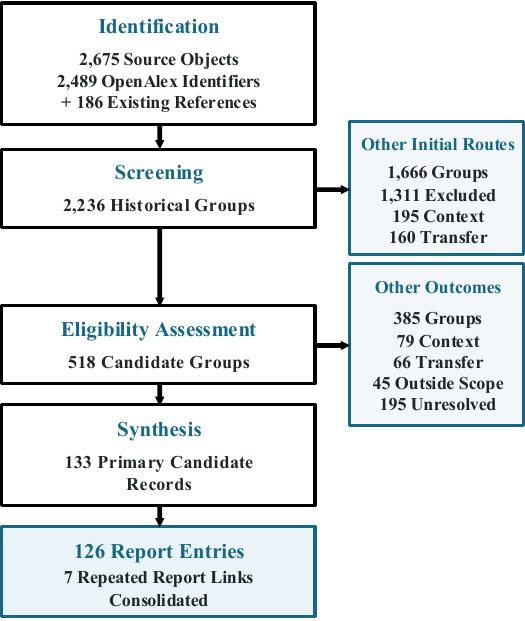}
\caption{\rev{PRISMA-informed literature selection. Counts distinguish source records, screening groups and report entries. Grouping precedes report-identity reconciliation; unresolved groups retain access or eligibility uncertainty. Detailed decisions and version links are retained in the review records.}}
\label{fig:review_flow}
\end{figure}

\rev{Figure~\ref{fig:review_flow} summarizes selection: 127 report entries are linked to 134 primary candidate records, with unresolved records identified separately. Retained review records document inspected sources, decision reasons, grouping corrections and report/version links. Counts remain report-level because related reports may share experiments.}

\subsection{Contributions}
This survey develops three contributions:
\begin{itemize}
    \item \textbf{Application taxonomy:} We organize LLM- and MM-LLM applications by operational domain and supporting functions, \rev{with actual model roles and input/output requirements as additional descriptors.}
    \item \textbf{Maturity-aware synthesis:} We distinguish operationally oriented applications, prototypes and simulation-based studies, and forward-looking concepts through source-specific comparisons and limitations. \rev{Validation maturity, publication status and action authority are recorded separately.}
    \item \textbf{Deployment-oriented agenda:} We synthesize recurring challenges related to latency, multi-modal alignment, explainability, governance, and evaluation, and translate them into research directions for practical TSMO integration.
\end{itemize}

\subsection{Organization of the Paper}
Section \ref{taxonomy} introduces LLM foundations, the TSMO context and related surveys. Section \ref{review} synthesizes applications. Sections \ref{gaps} and \ref{future} discuss deployment challenges and corresponding research directions, followed by conclusions in Section \ref{conclusions}.\par

\section{Foundations and Literature Synthesis}
\label{taxonomy}
\subsection{Large Language Models (LLMs)}
The development of LLMs has expanded natural language processing (NLP) capabilities \cite{Brown2020LanguageLearners,Zhao2023AModels}, building on longstanding NLP research \cite{Clark2010TheProcessing}. Transformer architectures use self-attention to represent contextual dependencies \cite{Vaswani2017AttentionNeed}.

Examples include the generative pre-trained transformer (GPT) series \cite{Brown2020LanguageLearners}, bidirectional encoder representations from transformers (BERT) \cite{Devlin2018BERT:Understanding}, LLaMA \cite{Touvron2023LLaMA:Models}, and Mamba \cite{Gu2023Mamba:Spaces}, spanning decoder, encoder and state-space designs. Pretraining learns language patterns from large corpora \cite{Touvron2023LlamaModels}; fine-tuning adapts models to domain-specific tasks \cite{Zhao2023AModels}.

\begin{figure}[!htbp]
\centering
\includegraphics[width=\linewidth]{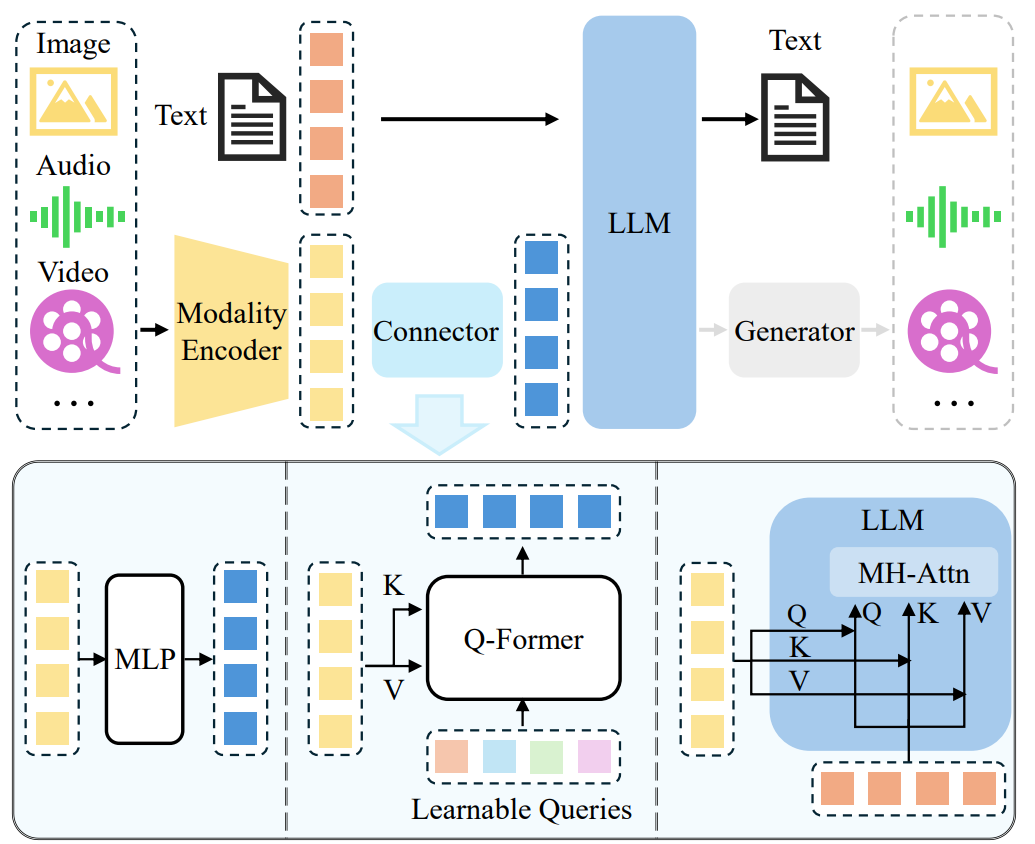}
\caption{An example of MM-LLM architecture \cite{Yin2023AModels}. MLP denotes multilayer perceptron; Q-Former, querying transformer; MH-Attn, multi-head attention; and Q, K, and V denote queries, keys, and values, respectively.}
\label{Fig:mm-llm}
\end{figure}

MM-LLMs integrate text with additional modalities, including images, video and sensor signals. Figure~\ref{Fig:mm-llm} shows an example architecture \cite{Yin2023AModels}. Recent surveys examine joint perception and reasoning \cite{Naveed2025AModels,Yin2023AModels}. Libraries such as LAVIS (i.e., A Library for Language-Vision Intelligence) provide supporting software infrastructure \cite{Li2022LAVIS:Intelligence}. Operational benefit and real-time suitability in TSMO require application-specific evaluation.

Computational cost, privacy and interpretability constrain deployment \cite{Naveed2025AModels}. Connections between transformers and state-space models offer complementary approaches to computational efficiency \cite{Dao2024TransformersDuality}. Applications include text summarization, classification, question answering and sentiment analysis.
Adaptation of LLM backbones can also support numerical forecasting \cite{Zhao2024EmbracingForecasting}, which is distinct from generating a linguistic explanation.

\rev{LLM-based transportation systems differ in how they receive information. Native MM-LLMs encode additional modalities directly, while hybrid systems supply an LLM with captions, serialized values or external tool results. Numerical adaptation of language backbones supports another class of predictive tasks. These interfaces shape the capabilities and evaluation requirements discussed in Section~\ref{review}.}

\subsection{Transportation Systems Management and Operations (TSMO)}
TSMO strategies optimize existing roads, vehicles and infrastructure to increase safety, reduce congestion and improve reliability without large-scale infrastructure expansion \cite{FederalHighwayAdministration2025TransportationOperations}. Transportation-systems analysis provides methods for assessing these objectives \cite{Casscetta2009TransportationApplications}.

Intelligent transportation systems (ITS) supply enabling technologies \cite{An2011ASystems,Dimitrakopoulos2010IntelligentFunctionality}, including real-time monitoring, dynamic message signs, global navigation satellite system (GNSS)-based vehicle tracking and traffic management centers (TMCs) \cite{Birriel2022ApplyingAreas}.

Within smart and connected communities, TSMO has expanded beyond roadway traffic control to include multi-modal integration, connected-vehicle data, traveler information, freight management, incident response, and infrastructure operations \cite{FederalHighwayAdministration2018TransportationCommunities}.

Data integration remains a significant hurdle because sources must be unified for operational use \cite{Du2024LargePerspective}. MM-LLMs offer mechanisms for processing textual, visual and sensor observations alongside numerical traffic data \cite{Mahmud2025IntegratingDirections}. Their actual input interfaces and measured timing determine suitability for time-sensitive TSMO tasks.

\subsection{Recent Surveys Related to This Topic}
Existing surveys and case studies cover public-transit and infrastructure information queries, routing guidance and decision support \cite{Jonnala2025ExploringStudy,Du2024LargePerspective}, while the San Antonio case illustrates retrieval from static transit feeds.

ITS surveys discuss model compression, hardware--software co-design and prompting for traffic management, demand forecasting and traveler information, alongside compute, privacy and benchmark constraints \cite{Mahmud2025IntegratingDirections,Wandelt2024LargeChallenges,Gan2024LargeSurvey,Shoaib2023ASystems}. Visual Mamba and multi-modal ITS surveys address architectures combining imagery, text and other inputs \cite{Xu2024VisualOutlooks,Le2024MultimodalSystems}. 

In automated driving, transformer-based language models support scene understanding and policy learning \cite{Zhao2023AModels,Yang2024LLM4Drive:Driving,Cui2023ADriving,Wang2025GenerativeOpportunities}, with related work on temporal consistency and hallucinations \cite{Dona2024EvaluatingTasks} and LLM-assisted Markov decision process specification for novice users \cite{Villarreal2023CanLearning}. Task-dependent results provide transferable context rather than deployed TSMO evidence. 

Forecasting studies combine numerical predictors with LLM representations \cite{Zhao2024EmbracingForecasting}; retrieval-augmented multi-agent work adds conversational interfaces for simulation, policy analysis and stakeholder engagement \cite{Xu2024GenAI-poweredSystems}. Operational value depends on integration and evaluation setting. 

\rev{Existing surveys distinguish decision making from guidance \cite{Nie2025ExploringRoadmap}, compare traffic models and input--output interfaces \cite{Yan2025LargeOpportunities}, and examine autonomy, human roles and failure consequences \cite{Tu2026IntegrationBoundaries}. Table~\ref{tab:novelty} positions this survey's contribution, linking system-level TSMO tasks to actual model functions, validation evidence and operational requirements.}
\begin{table*}[!tbp]\reviewfloatcolor
\centering
\changedcolor
\caption{\rev{Coverage in previous LLM-based transportation surveys.}}
\label{tab:novelty}
\footnotesize
\setlength{\tabcolsep}{3pt}
\renewcommand{\arraystretch}{1.10}
\begin{tabularx}{\textwidth}{@{}>{\raggedright\arraybackslash}p{1.8cm}c>{\raggedright\arraybackslash}X*{5}{>{\centering\arraybackslash}p{0.09\textwidth}}@{}}
\toprule
\textbf{Survey} & \textbf{Year} & \textbf{Primary Focus} & \textbf{System-level Applications} & \textbf{Model Roles / Interfaces} & \textbf{Quantitative Comparisons} & \textbf{Validation / Deployment} & \textbf{Operational Assurance} \\
\midrule
Wandelt et al.~\cite{Wandelt2024LargeChallenges} & 2024 & Transportation applications and implementation challenges & \checkmark & (\checkmark) & ? & (\checkmark) & (\checkmark) \\
\addlinespace[4pt]
Mahmud et al.~\cite{Mahmud2025IntegratingDirections} & 2025 & Intelligent transportation use cases, architectures and challenges & \checkmark & \checkmark & \checkmark & (\checkmark) & \checkmark \\
\addlinespace[4pt]
Yan et al.~\cite{Yan2025LargeOpportunities} & 2025 & Preprint: transportation applications, interfaces and methods & \checkmark & \checkmark & \checkmark & (\checkmark) & (\checkmark) \\
\addlinespace[4pt]
Nie et al.~\cite{Nie2025ExploringRoadmap} & 2025 & Role-based transportation framework and practical guidance & \checkmark & \checkmark & \checkmark & (\checkmark) & \checkmark \\
\addlinespace[4pt]
Tu et al.~\cite{Tu2026IntegrationBoundaries} & 2026 & Integration depth, capability boundaries and hybrid traffic intelligence & \checkmark & \checkmark & (\checkmark) & \checkmark & \checkmark \\
\midrule
\textbf{This Survey} & 2026 & TSMO tasks linked to model functions, validation evidence and operational requirements & \checkmark & \checkmark & \checkmark & \checkmark & \checkmark \\
\bottomrule
\end{tabularx}
\par\vspace{5pt}\noindent
\begin{minipage}{\textwidth}
\footnotesize
\checkmark: explicit comparative treatment; (\checkmark): discussion or selected examples; ?: insufficient accessible evidence. 
\end{minipage}
\end{table*}

\section{Applications of LLMs in TSMO}
\label{review}
\begin{figure*}[!tbp]\reviewfloatcolor
\centering
\includegraphics[width=\linewidth]{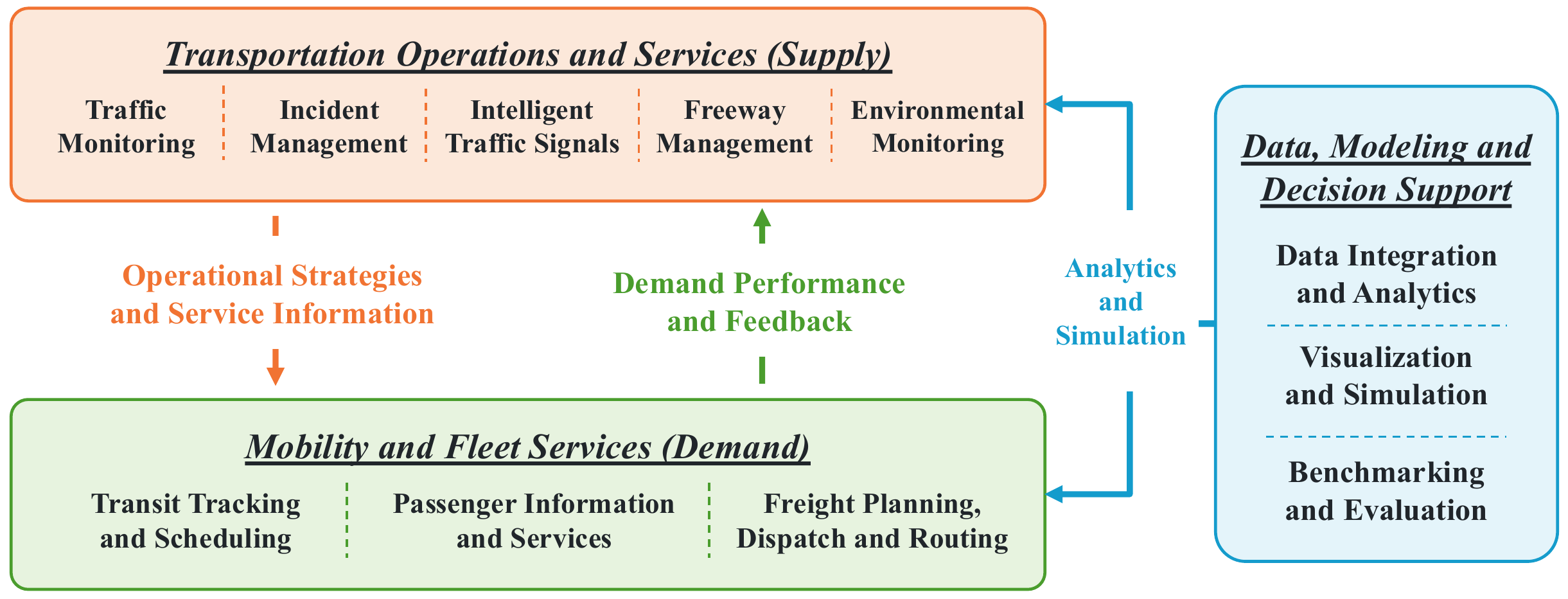}
\caption{\rev{Application organization of the survey. Transportation operations and mobility services share a supporting layer for data, modeling and decision support.}}
\label{Fig:taxonomy}
\end{figure*}

\rev{This review organizes system-level TSMO applications into two operational domains and a shared supporting layer:}
\begin{itemize}
    \item \textbf{Transportation Operations \& Services (Supply):} Traffic monitoring, incident management, intelligent signals, freeway management and environmental monitoring.
    \item \textbf{Mobility \& Fleet Services (Demand):} Transit tracking and scheduling, passenger information and services, and freight planning, dispatch and routing.
    \item \textbf{Data, Modeling \& Decision Support:} Data integration, visualization, simulation and evaluation functions supporting both operational domains.
\end{itemize}
Figure~\ref{Fig:taxonomy} summarizes this organization. Table~\ref{tab:operational_io} connects the applications to their inputs, output users and verification requirements. \rev{A report may address several functions while retaining one identity in the review records.}

\rev{The predominantly offline and simulated evidence in Table~\ref{tab:maturity_report_counts} motivates separate assessment of task performance, operational integration and LLM authority.}

\subsection{Transportation Operations \& Services (Supply)}
\label{sub:supply}
Monitoring informs incident response, which drives adaptive signal control; roadway and environmental inputs adjust both detection and control strategies. LLMs can serve as integrators by turning heterogeneous text and sensor-derived information into unified, context-aware recommendations.

\subsubsection{Traffic Monitoring}
\label{sub:monitoring}
Real-time traffic monitoring has traditionally relied on loop detectors, radar sensors, and camera networks to measure vehicle counts, speeds, and occupancies. However, sensor-only systems often miss contextual nuances, such as localized hazards, emerging weather effects, special events, or traveler-reported disruptions. LLM-based tools can ingest unstructured updates and extract operationally relevant details for traffic managers \cite{Da2024Open-ti:Model}. Social media data have also been used to identify traffic-accident occurrence and severity \cite{Ali2021TrafficData}, while earlier NLP systems interpret colloquial descriptions of road conditions \cite{Wan2020EmpoweringData}. Dedicated traffic-management chatbots, including TP-GPT, allow operators to query transportation databases in natural language \cite{Wang2024TrafficManagement}. Related work combines physics-informed traffic estimation with language-based traveler assistance during disruptions \cite{SyumGebre2024AI-IntegratedAssistance}.

Tables~\ref{tab:supply_overview} and~\ref{tab:traffic} provide an overview and selected comparisons across these supply-side applications.

Query assistance and visual interpretation offer complementary capabilities. \rev{Structured prompting and checking improve database-query results, but executable queries can still return imperfect answers \cite{Wang2024TrafficManagement}.} \rev{Reusing information across overlapping cameras reduces repeated video-to-text processing \cite{Arefeen2025TrafficLens}. \phantomsection\label{par:tau_edge_evidence}Edge-device anomaly classification and summarization further demonstrate a practical execution path \cite{Lin2026TAUR1}. Across these tasks, correctness, processing time and coverage of unfamiliar sites require separate assessment; faster ingestion alone is insufficient evidence of faster incident detection.}

Beyond chatbot-based responses, LLM-inspired architectures support forecasting through learned traffic representations \cite{Moghadas2025Strada-LLM:Prediction}. Textual traffic, weather and geographic context can enrich forecasts \cite{Guo2024TowardsModels}, while numerical backbone adaptation encodes spatial-temporal traffic patterns directly \cite{Liu2024Spatial-TemporalPrediction}. Tool-based systems support visualization and interpretation of traffic states \cite{Zhang2024TrafficGPT:Models}, and text-image instruction tuning supports transportation question answering \cite{Wang2024TransGPT:Transportation}. \rev{These roles should be distinguished: improvements vary by error metric and prediction horizon, generated explanations require their own grounding checks, and an inspired architecture alone does not establish the use of language-pretrained weights. Matched data, context and baselines are therefore essential to forecasting comparisons.}

\begin{table*}[!tbp]\reviewfloatcolor
\centering
\changedcolor
\caption{\rev{Operational inputs, outputs and verification requirements for the application review.}}
\label{tab:operational_io}
\footnotesize
\setlength{\tabcolsep}{4pt}
\renewcommand{\arraystretch}{1.10}
\begin{tabularx}{\textwidth}{@{}p{2.5cm} >{\raggedright\arraybackslash}X >{\raggedright\arraybackslash}X >{\raggedright\arraybackslash}X@{}}
\toprule
\textbf{Operational Task} & \textbf{Required Input and Representation} & \textbf{Output and Decision Consumer} & \textbf{Independent Check Before Use} \\
\midrule
Network Monitoring / Querying & Time-stamped detector values, camera-derived observations, network identifiers, database schema and operator question & Source-linked state summary or query result for a TMC analyst & Query permissions, location/time alignment, numerical consistency and source freshness \\
\addlinespace[3pt]
Incident Coordination & Ticket narrative, verified location/direction, event state, response procedures and available resources & Structured incident facts and proposed response for an incident commander & Missing critical facts, procedure compliance, resource feasibility and explicit approval \\
\addlinespace[3pt]
Signal / Corridor Control & Lane/phase definitions, queues or arrivals, controller state and conflict/timing constraints & Phase or timing proposal for a controller/operator & Feasible phase, minimum timing, conflict exclusion, deadline and approved fallback \\
\addlinespace[3pt]
Transit / Traveler Services & Versioned feeds, disruption alerts, accessibility requirements and passenger preferences & Itinerary or service advisory for staff/riders & Connectivity, alert validity, accessibility constraints and publishing authorization \\
\addlinespace[3pt]
Freight / Fleet Operations & Orders, depots, vehicles, capacities, time windows and road restrictions & Demand estimate or routing/dispatch proposal for a fleet planner & Forecast target consistency; solver feasibility and cost under actual constraints \\
\addlinespace[3pt]
Simulation / Evaluation & Versioned network, demand, events, model parameters and requested experiment & Executable scenario, calibrated model or evaluation report for an analyst & Syntax plus physical plausibility, calibration, reproducibility and stated simulation assumptions \\
\bottomrule
\end{tabularx}
\end{table*}

Taken together, the literature supports an operator-assistive role for information fusion, natural-language querying and forecasting support \cite{Yan2025LargeOpportunities}. The main integration challenge is to preserve source accuracy and freshness as information moves from observation to interpretation and operator use.

\subsubsection{Incident Management}
\label{sub:incidents}
When a crash or other disruption occurs, operators may receive incident tickets, crash narratives, connected-vehicle messages, camera observations, and field reports within a short time window. LLMs can help summarize, classify, and prioritize this information. Structured prompting has been explored for crash-severity analysis \cite{Zhen2024LeveragingInference}, while hybrid pipelines convert accident descriptions into features for conventional classifiers \cite{Grigorev2024EnhancingClassification}. Generative incident-response systems suggest action plans from reports and response guidance \cite{Grigorev2024IncidentResponseGPT:Intelligence}. Highway incident-management studies also explore accessible, data-driven decision-support workflows \cite{Cercola2025AutomatingHighway}.

Multi-modal incident analysis extends this capability by combining language with perception and connected-vehicle information \cite{Wang2023AccidentGPT:Model}. Domain adaptation uses transportation safety materials and historical crash records to support specialized assessment \cite{Zheng2023TrafficSafetyGPT:Safety,Zhao2025SafeTrafficInterventions}. \phantomsection\label{par:severity_fusion_evidence}\rev{Comparisons of semantic-feature integration show that gains depend on how language-derived information is combined with existing predictors; simple concatenation can perform worse than structured data alone \cite{Hao2026SeveritySemanticFeatures}. This favors evaluating the added information and integration strategy together, with attention to rare severe events and data-access constraints.}

\rev{The evidence is strongest for retrospective classification and structured plan assistance. Reference-plan agreement, text similarity and predictive attribution answer different questions from clearance time, dispatch effectiveness or the causal effect of an intervention. These distinctions support a workflow in which language models organize evidence and propose responses, while operators check actions against current conditions and agency procedures.}

\begingroup
\renewcommand{\arraystretch}{1.10}
\begin{table*}[!tbp]\reviewfloatcolor
\centering
\caption{\rev{Summary of LLM applications in transportation operations \& services (Supply).}}
\label{tab:supply_overview}
\footnotesize
\setlength{\tabcolsep}{4pt}
\begin{tabular}{@{}
>{\raggedright\arraybackslash}p{\dimexpr0.16\textwidth-1.5\tabcolsep\relax}
>{\raggedright\arraybackslash}p{\dimexpr0.22\textwidth-1.5\tabcolsep\relax}
>{\raggedright\arraybackslash}p{\dimexpr0.28\textwidth-1.5\tabcolsep\relax}
>{\raggedright\arraybackslash}p{\dimexpr0.34\textwidth-1.5\tabcolsep\relax}
@{}}
\toprule
\textbf{Category} & \textbf{Key References} & \textbf{Applications} & \textbf{LLM Role \& Evidence Boundary} \\
\midrule
\textbf{Traffic Monitoring}
& \cite{Ali2021TrafficData,Wan2020EmpoweringData} (precursors);
\cite{Zhang2024TrafficGPT:Models,Liu2024Spatial-TemporalPrediction};
\cite{Wang2024TransGPT:Transportation} (domain-model context);
\cite{SyumGebre2024AI-IntegratedAssistance}
& Event extraction; data queries/visualization; numerical prediction; traffic-state assistance.
& Queries/numerical features; precursor NLP and domain question answering provide context. State estimates use a physics-informed neural network; operational timing unverified. \\
\addlinespace[3pt]
\textbf{Incident Management}
& \cite{Zhen2024LeveragingInference,Grigorev2024IncidentResponseGPT:Intelligence,Wang2023AccidentGPT:Model,Zheng2023TrafficSafetyGPT:Safety};
\cite{Somvanshi2024Gen-AIManagement} (framework context)
& Severity classification; response plans; simulated accident analysis; safety question answering.
& Domain question answering and planning; retrospective/simulation evidence. \\
\addlinespace[3pt]
\textbf{Intelligent Signals}
& \cite{Masri2024LeveragingScenarios,Pang2024ILLM-TSC:Improvement,Lai2024LLMLight:Agents}
& Adaptive phases; RL--LLM control; mixed-traffic conflict advice.
& Simulated phase selection/RL overrides; speed/lane/yielding advice is adjacent evidence. Field robustness needs evaluation. \\
\addlinespace[3pt]
\textbf{Freeway Management}
& \cite{Wang2017ExploringRetrieval} (precursor context); \cite{Tong2025RoadSituationCITS}
& Accident localization; highway-image interpretation; standardized warnings.
& Spatial retrieval/message generation; lane-status accuracy and live delivery need validation. \\
\addlinespace[3pt]
\textbf{Environmental Monitoring}
& \cite{Marathe2023WEDGE:Models};
\cite{Yin2024CrisisSense-LLM:Informatics} (transfer evidence); \cite{Guo2024TowardsModels,Da2024PromptLearning}
& Synthetic weather images; disaster-text classification; weather-conditioned forecasts; simulated dynamics specification.
& Forecasting and simulator parameters; imagery, text and simulation evidence. \\
\bottomrule
\end{tabular}
\end{table*}
\endgroup

\begin{table*}[!tbp]\reviewfloatcolor
\centering
\changedcolor
\caption{\rev{Selected quantitative comparisons for traffic monitoring and signal control.}}
\label{tab:traffic}
\footnotesize
\setlength{\tabcolsep}{4pt}
\renewcommand{\arraystretch}{1.10}
\begin{tabularx}{\textwidth}{@{}>{\raggedright\arraybackslash}p{0.20\textwidth} >{\raggedright\arraybackslash}p{0.24\textwidth} >{\raggedright\arraybackslash}X >{\raggedright\arraybackslash}X@{}}
\toprule
\textbf{Application / Study} & \textbf{Task and Evaluation Setting} & \textbf{Key Findings} & \textbf{Operational Implications and Limitations} \\
\midrule
\textbf{Traffic Monitoring}\newline TP-GPT \cite{Wang2024TrafficManagement} & Database queries; 50 TransQuery tasks (M2) & Flawless answers: 80\% versus 44\% for GPT-4 Turbo & Query checking improves this test; executable queries still require answer validation. \\
\addlinespace[3pt]
\textbf{Traffic Monitoring}\newline ST-LLM \cite{Liu2024Spatial-TemporalPrediction} & Taxi/bike-flow prediction; mean over 12 half-hour horizons (M2) & Taxi-pickup MAE: 5.29 versus GWN's 5.43 & RMSE slightly favors GWN. Numerical prediction is distinct from linguistic explanation. \\
\addlinespace[3pt]
\textbf{Intelligent Signals}\newline LLMLight / LightGPT \cite{Lai2024LLMLight:Agents} & Signal selection in CityFlow; Hangzhou-2 (M2) & Average waiting time: 64.16 versus 76.59 s for Advanced-CoLight & Waiting-time gains coexist with higher travel time on the same test. \\
\addlinespace[3pt]
\textbf{Intelligent Signals}\newline VLMLight \cite{Wang2025VLMLight:Reasoning} & Emergency priority; three SUMO intersections (M2) & Songdo emergency waiting: 7.48 versus 22.04 s for UniTSA & Ordinary-traffic travel time rises; legality review remains a simulator safeguard. \\
\addlinespace[3pt]
\textbf{Intelligent Signals}\newline Traffic-R1 \cite{Zou2025Traffic-R1:Systems} & Ten intersections; six-week alternating-week trial, weekday evening peaks (M3, author-reported) & Mean queue: 31.3 versus 34.5 under manual control; unit unreported & Expert supervision; absent demand matching and uncertainty estimates limit attribution. \\
\bottomrule
\end{tabularx}
\par\vspace{3pt}\begin{minipage}{\textwidth}\footnotesize MAE: mean absolute error; RMSE: root mean squared error; GWN: Graph WaveNet; SUMO: Simulation of Urban MObility.\end{minipage}
\end{table*}

Local policy and operational language are especially important for this application. Regulation-aware retrieval and policy adaptation in automated driving provide transferable approaches to domain grounding \cite{Cai2025DrivingLLM,Li2024DrivingAdaptation}. An MM-LLM-based agent could populate incident-management records with location, incident type and severity, then assist with response planning and public information. Generative AI (Gen-AI) frameworks for TSMO knowledge management identify this coordination role for major crashes and multi-incident scenarios \cite{Somvanshi2024Gen-AIManagement}.

\subsubsection{Intelligent Traffic Signals}
\label{sub:signals}
Traditional signal optimization often relies on control algorithms or reinforcement learning (RL) driven primarily by sensor measurements. Where an evaluated controller receives only numerical states, narrative reports and changing operational constraints require an additional interface. LLM-assisted intersection-management studies explore how high-level descriptions and constraints can inform decisions \cite{Masri2024LeveragingScenarios}. Hybrid controllers allow an LLM to review an RL-based phase decision \cite{Pang2024ILLM-TSC:Improvement}.

LLMs may also contribute explanation and design assistance to intersection operations. \rev{Assistance with observation and reward functions supports controller development in mixed-traffic simulations \cite{Villarreal2023CanLearning}, whereas phase-selection agents participate directly in simulated signal control \cite{Lai2024LLMLight:Agents}. Their evaluations show task-dependent gains: improved waiting time can coexist with higher overall travel time, and assistance that benefits one control problem may not benefit another. The distinction between design-time support and runtime decisions therefore matters when interpreting performance.}

\rev{Priority handling makes these trade-offs particularly visible. Combining visual interpretation and deliberative phase selection can reduce emergency-vehicle waiting while imposing ordinary-traffic penalties \cite{Wang2025VLMLight:Reasoning}. \phantomsection\label{par:signal_event_baseline}Event-aware comparisons also show that generated control skills can improve average delay yet respond less effectively to an emergency than matched handcrafted preemption \cite{Lei2026SignalClaw}. This supports evaluating both network efficiency and priority objectives, rather than relying on a single aggregate score. Offline skill generation and inference within a control cycle also have different timing requirements.}

\rev{Operational evidence includes an author-reported, supervised six-week pilot at ten arterial intersections \cite{Zou2025Traffic-R1:Systems}. Routine plans required expert approval and incident adjustments remained supervised. Reported queue reductions provide bounded pilot evidence, with demand matching, measurement units and independent corroboration still limiting causal interpretation.}

Weather-conditioned policy transfer offers another use of language-derived context, although the inspected experiments transfer between simulators \cite{Da2024PromptLearning}. \rev{Across signal studies, common dataset names do not ensure comparable training regimes, control intervals or permitted actions. Matched controllers, fallback behavior and explicit priority constraints are more informative than broad claims of model superiority.}

\subsubsection{Freeway Management}
\label{sub:freeway}
LLMs can triage maintenance logs, work orders and citizen reports about road conditions. Natural-language location extraction provides a foundation for converting incident descriptions into usable spatial information \cite{Wang2017ExploringRetrieval}. Within the TSMO context, language models could extend this capability by jointly interpreting citizen reports, maintenance logs and sensor anomalies to support issue triage and maintenance prioritization. \phantomsection\label{par:emergency_comparator}\rev{Existing congestion-prediction and emergency-lane control methods provide relevant operational comparators \cite{Tang2026EmergencyLane}; evaluations should identify the incremental contribution of language-based interpretation within those workflows.}

\phantomsection\label{par:cits_evidence}\rev{Standardized warning generation connects image interpretation and parameter extraction to cooperative intelligent transport systems (C-ITS) message encoding \cite{Tong2025RoadSituationCITS}. The contrast between high schema correctness and weak lane-status identification shows why both message structure and operational meaning must be checked. Request-level timing is only one part of a warning pipeline; transmission, review and vehicle response also determine its operational usefulness.}

\subsubsection{Environmental Monitoring}
\label{sub:environment}
During extreme events, operators may receive weather advisories, field reports, social media posts, and road-condition updates simultaneously. LLMs and MM-LLMs can help consolidate these information streams into actionable operational insights. Vision-language-generated datasets such as WEDGE support perception testing under diverse weather conditions \cite{Marathe2023WEDGE:Models}. Physics-informed estimation with language-based assistance addresses disruption interpretation \cite{SyumGebre2024AI-IntegratedAssistance}, while disaster-text classification organizes social media reports during emergency events \cite{Yin2024CrisisSense-LLM:Informatics}.

\rev{These applications connect numerical forecasts, visual observations and unstructured reports, but evaluate different parts of an environmental-response workflow. Perception robustness, information classification and weather-conditioned control should therefore be assessed separately. Timestamp and spatial alignment, false alarms and operator escalation are common requirements for translating the resulting information into useful action.}

\subsection{Mobility \& Fleet Services (Demand)}
\label{sub:demand}
LLM-empowered interfaces can support vehicle tracking and scheduling workflows, deliver conversational passenger notifications and support, and assist prompt-driven multi-modal planning. Tables~\ref{tab:demand_overview} and~\ref{tab:public} provide an overview and selected comparisons across mobility and fleet-service applications.

\begingroup
\renewcommand{\arraystretch}{1.10}
\begin{table*}[!tbp]\reviewfloatcolor
\centering
\caption{\rev{Summary of LLM applications in mobility \& fleet services (Demand).}}
\label{tab:demand_overview}
\footnotesize
\setlength{\tabcolsep}{4pt}
\begin{tabular}{@{}
>{\raggedright\arraybackslash}p{\dimexpr0.16\textwidth-1.5\tabcolsep\relax}
>{\raggedright\arraybackslash}p{\dimexpr0.22\textwidth-1.5\tabcolsep\relax}
>{\raggedright\arraybackslash}p{\dimexpr0.28\textwidth-1.5\tabcolsep\relax}
>{\raggedright\arraybackslash}p{\dimexpr0.34\textwidth-1.5\tabcolsep\relax}
@{}}
\toprule
\textbf{Category} & \textbf{Key References} & \textbf{Applications} & \textbf{LLM Role \& Evidence Boundary} \\
\midrule
\textbf{Transit Tracking \& Scheduling}
& \cite{Chen2024DelayPTC-LLM:Models,Fang2024TraveLLM:Disruption,Ying2024BeyondPlanning,Devunuri2024TransitGPT:Models}
& Delay-related passenger-choice prediction; disruption alternatives; general transit feed specification (GTFS) queries/code.
& Static/retrospective prompting and data access; live tracking/schedule interventions untested. \\
\addlinespace[3pt]
\textbf{Passenger Information \& Services}
& \cite{Jonnala2025ExploringStudy,Devunuri2024TransitGPT:Models}
& Transit queries; service-information retrieval/presentation.
& Static-feed language interfaces; rider satisfaction, booking and live-alert outcomes unmeasured. \\
\addlinespace[3pt]
\textbf{Freight Planning, Dispatch \& Routing}
& \cite{Tupayachi2024TowardsTransportation,Nie2025JointApproach};
\cite{Lu2025Fine-TuningPlatforms} (pricing transfer);
\cite{Felder2025SmartApproach,Xu2025TowardsProtocol};
\cite{Liu2023CanOptimization} (precursor context);
\cite{HuangFromRouting,Li2025ARS:Models}
& Inter-modal ontology; delivery demand/pricing; planning; route/solver generation.
& Embeddings/solver generation; Word2Vec and pricing are context. Dispatch/emissions benefits untested. \\
\bottomrule
\end{tabular}
\end{table*}
\endgroup

\begin{table*}[!tbp]\reviewfloatcolor
\centering
\changedcolor
\caption{\rev{Selected comparisons within the Demand domain, spanning transit information, disruption response and freight forecasting/routing.}}
\label{tab:public}
\footnotesize
\setlength{\tabcolsep}{4pt}
\renewcommand{\arraystretch}{1.10}
\begin{tabularx}{\textwidth}{@{}>{\raggedright\arraybackslash}p{0.20\textwidth} >{\raggedright\arraybackslash}p{0.24\textwidth} >{\raggedright\arraybackslash}X >{\raggedright\arraybackslash}X@{}}
\toprule
\textbf{Application / Study} & \textbf{Task and Evaluation Setting} & \textbf{Key Findings} & \textbf{Operational Implications and Limitations} \\
\midrule
\textbf{Transit Tracking \& Scheduling}\newline TransitGPT \cite{Devunuri2024TransitGPT:Models} & Static GTFS queries; 100 tasks (M2) & Examples and error feedback raise GPT-4o accuracy from 74\% to 90\% & Timing effects vary by model; real-time tracking remains untested. \\
\addlinespace[3pt]
\textbf{Transit Tracking \& Scheduling}\newline DelayPTC-LLM \cite{Chen2024DelayPTC-LLM:Models} & Retrospective metro waiting/abandonment prediction (M2) & Accuracy: 0.83 versus GPT-4's 0.53, but lower reported F1 & Metric disagreement and unresolved F1 reporting limit interpretation. \\
\addlinespace[3pt]
\textbf{Transit Tracking \& Scheduling}\newline TraveLLM \cite{Fang2024TraveLLM:Disruption} & Disruption-aware routing; nine designed scenarios (M2) & Connectivity score: GPT-4 0.78 versus Gemini 0.11 & Designed scenarios reveal route-feasibility failures; passenger time savings require separate evaluation. \\
\addlinespace[3pt]
\textbf{Passenger Information \& Services}\newline TransitTalk predecessor \cite{Wang2024LeveragingServices} & Alert, itinerary and policy assistance; 2024 preprint prototypes (M1) & Demonstrated language-assisted service-information workflows & Human review before alerts; later published implementation remains uninspected. \\
\addlinespace[3pt]
\textbf{Freight Planning, Dispatch \& Routing}\newline IMPEL \cite{Nie2025JointApproach} & Geographic embeddings with delivery-demand prediction (M2) & Shanghai MAE: 3.76 versus SATCN's 4.75 & Forecasting evidence; some data are proprietary and dispatch gains are untested. \\
\addlinespace[3pt]
\textbf{Freight Planning, Dispatch \& Routing}\newline ARS \cite{Li2025ARS:Models} & Routing-program generation on hard RoutBench variants (M2) & Program success: 46.80\% versus Reflexion's 15.20\% & Program success requires valid routes. Optimality and runtime remain important limitations. \\
\bottomrule
\end{tabularx}
\par\vspace{4pt}\noindent\begin{minipage}{\textwidth}\footnotesize SATCN: spatial aggregation and temporal convolutional network; MAE: mean absolute error; F1: precision--recall score.\end{minipage}
\end{table*}

\phantomsection\label{par:fleet_service_evidence}\rev{Fleet-service workflows distinguish optimization assistance from simulated dispatch. Translating operator objectives into solver constraints can reduce optimization effort \cite{Jiang2025RideAgent}, while learned LLM policies support order matching and driver repositioning in simulation \cite{Lyu2026LLMODDR}. These approaches should be compared through solution quality, computation and service performance under the same operational constraints.}

\subsubsection{Transit Tracking \& Scheduling}
\label{sub:transit}
Public transit systems produce vast operational text, such as driver logs, passenger complaints and service bulletins, alongside GNSS-based vehicle location feeds. LLMs can assist by correlating numeric data with these textual sources to provide a fuller picture of service conditions. Delay-related studies predict whether metro passengers wait or abandon a trip after receiving disruption information \cite{Chen2024DelayPTC-LLM:Models}. This provides passenger-response information for planning, distinct from directly changing a timetable or holding a bus.

LLMs are also being integrated with transit operations planning and analysis. In public-transit disruption scenarios, language models can assist with route planning by interpreting user goals, network constraints and disruption descriptions \cite{Fang2024TraveLLM:Disruption}. Broader planning evaluations examine congestion-pricing guidance and simulation or tabular analysis \cite{Ying2024BeyondPlanning}. \rev{Across these tasks, an apparently useful recommendation must still satisfy connectivity and disruption constraints. Differences between accuracy and F1-score in passenger-choice prediction further highlight the importance of evaluating model performance using multiple complementary metrics.}

Transit agencies typically have troves of data, such as schedules, real-time feeds and performance metrics, that are not easily accessible except through specialized tools. Natural-language interfaces can lower this access barrier by translating questions into code or database queries and returning the result \cite{Devunuri2024TransitGPT:Models}. \rev{Current static general transit feed specification (GTFS) evaluations support timetable lookup and transfer calculations; live tracking requires the additional integration and validation of real-time feeds.}

\subsubsection{Passenger Information \& Services}
\label{sub:passenger}
Conversational LLM interfaces can support rider questions, service-change and disruption messages, issue reporting and multilingual interaction while accommodating varied natural-language requests. 

Practitioner examples illustrate different stages of adoption. Visual parking-sign interpretation provides related traveler-information context \cite{Arafat2024ParkingGPTCopilot}. The International Association of Public Transport (UITP) report describes a planned LLM upgrade for the Chicago Transit Authority (CTA) service Chat with CTA and a cloud-based LLM announcement pipeline at Swiss PostBus \cite{AdvancingPublicTransportUITP2025ArtificialTransport}. Static transit-data studies examine the accuracy of answering service questions \cite{Jonnala2024UsingStudy}. \rev{Together, they motivate accessible information services while keeping planned functions, tested queries and operational reporting distinct.}

\rev{Alert writing, itinerary assistance and policy explanation are illustrated in the accessible TransitTalk predecessor, which requires human review before publishing alerts \cite{Wang2024LeveragingServices}. Evidence from that prototype is kept separate from its published follow-on \cite{Wang2025TransitTalkAssist}. For public-facing use, source freshness, accessibility information and review of safety-sensitive advice are central safeguards.}

\subsubsection{Freight Planning, Dispatch \& Routing}
\label{sub:freight}
Port berth allocation provides related logistics context \cite{Wang2025IntelligentAllocation}. Ontology and knowledge-graph workflows convert scientific articles, technical manuals and scenario descriptions into representations for inter-modal planning and data integration \cite{Tupayachi2024TowardsTransportation}. Agentic digital-twin frameworks further coordinate optimization and simulation tools \cite{Xu2025TowardsProtocol}. \rev{Their main contribution is organizing knowledge and analytical workflows; competency-query performance and demonstration feasibility should be distinguished from transport outcomes.}

A closely related line of research addresses freight demand estimation and market-facing prediction. Geographic LLM embeddings enrich graph-based estimation and prediction of city-wide delivery demand, including cross-city transfer \cite{Nie2025JointApproach}. Fine-tuning has also been explored for network freight-price prediction in limited-data settings \cite{Lu2025Fine-TuningPlatforms}. \rev{These applications show how contextual representations can complement structured predictors, with demand estimation and market pricing serving different operational consumers.}

Routing and dispatch are active directions for freight-oriented LLM applications. Knowledge graphs and graph optimization support route enrichment and completion \cite{Felder2025SmartApproach}; route-sequence representations learned from driver behavior provide a distinct precursor approach \cite{Liu2023CanOptimization}. Generative models translate natural-language constraints into routing programs through debugging and verification \cite{HuangFromRouting,Li2025ARS:Models}. \rev{Related work addresses time-window routing \cite{Cao2025AWindows} and energy-aware formulations, with the latter's implementation still subject to full-text access \cite{Han2025TheRoads}. General optimization-modeling frameworks provide further methodological context \cite{Huang2024ORLM:Modeling}.}

\rev{Constraint adaptability is a clear shared motivation, but it involves a trade-off with specialist solver efficiency. Routing-program generation improves validated-solution success over general prompting methods, yet difficult variants retain substantial failures and conventional solvers can be much faster on standard problems \cite{Li2025ARS:Models}. A useful freight workflow therefore combines flexible problem specification with independent feasibility checks and comparison of costs under actual fleet constraints.}

\begingroup
\renewcommand{\arraystretch}{1.10}
\begin{table*}[!tbp]\reviewfloatcolor
\centering
\caption{\rev{Summary of LLM applications in data, modeling, evaluation \& decision support.}}
\label{tab:support_overview}
\footnotesize
\setlength{\tabcolsep}{4pt}
\begin{tabular}{@{}
>{\raggedright\arraybackslash}p{\dimexpr0.16\textwidth-1.5\tabcolsep\relax}
>{\raggedright\arraybackslash}p{\dimexpr0.22\textwidth-1.5\tabcolsep\relax}
>{\raggedright\arraybackslash}p{\dimexpr0.28\textwidth-1.5\tabcolsep\relax}
>{\raggedright\arraybackslash}p{\dimexpr0.34\textwidth-1.5\tabcolsep\relax}
@{}}
\toprule
\textbf{Category} & \textbf{Key References} & \textbf{Applications} & \textbf{LLM Role \& Evidence Boundary} \\
\midrule
\textbf{Data Integration \& Analytics}
& \cite{Somvanshi2024Gen-AIManagement} (framework context);
\cite{Jonnala2025ExploringStudy};
\cite{Yan2025LargeOpportunities} (survey context); \cite{Huang2024ORLM:Modeling} (optimization-method context)
& Knowledge management; transit interpretation; retrieval; optimization-model construction.
& Knowledge/query interfaces and static GTFS analysis; surveys/general optimization provide context. \\
\addlinespace[3pt]
\textbf{Visualization \& Simulation Platforms}
& \cite{Chang2024LLMScenario:Generation}; \cite{Guzay2023AGeneration} (details unverified);
\cite{Ye2025SUMO-MCP:Optimization,Wei2024EditableLLM-Agents,Zhong2023Language-GuidedDiffusion,Liu2024ControllableReasoning,Da2024PromptLearning,Wang2024LargeGeneration}
& Language-guided scene generation/editing; simulator calls; signal-policy transfer; personal mobility generation.
& Simulator interfaces; autonomous-driving scenes/individual activities are transfer/context. Inaccessible details unverified. \\
\addlinespace[3pt]
\textbf{Benchmarking \& Evaluation Platforms}
& \cite{Syed2024BenchmarkingBehaviors,Kwon2025TrafficNetQA:Files,Chaudhuri2025TripCraft:Planning,Song2026MobilityBench:Scenarios,Zhang2024TransportationGames:Models}
& Transportation knowledge/network queries; constrained itineraries; route-agent tests.
& Offline knowledge, constraints and tool-use tests; reproducibility depends on data/protocol access. \\
\bottomrule
\end{tabular}
\end{table*}
\endgroup

\subsection{Data, Modeling \& Decision Support}
\label{sub:support}
Table \ref{tab:support_overview} outlines the foundational techniques that enable and enhance TSMO applications. \rev{This supporting layer enables both supply- and demand-side applications rather than forming an additional mutually exclusive operational domain.}

\begin{table*}[!tbp]\reviewfloatcolor
\centering
\changedcolor
\caption{\rev{Selected evaluation resources. Knowledge tests and operational task benchmarks have different assessment boundaries.}}
\label{tab:eval}
\footnotesize
\setlength{\tabcolsep}{4pt}
\renewcommand{\arraystretch}{1.10}
\begin{tabularx}{\textwidth}{@{}>{\raggedright\arraybackslash}p{0.20\textwidth} >{\raggedright\arraybackslash}p{0.24\textwidth} >{\raggedright\arraybackslash}X >{\raggedright\arraybackslash}X@{}}
\toprule
\textbf{Application / Study} & \textbf{Task and Evaluation Setting} & \textbf{Key Findings} & \textbf{Operational Implications and Limitations} \\
\midrule
\textbf{Engineering Knowledge}\newline TransportBench \cite{Syed2024BenchmarkingBehaviors} & Curated transportation problems; context benchmark & Measures answer accuracy, consistency and reasoning behavior & Evaluation covers domain knowledge; control actions require separate testing. \\
\addlinespace[3pt]
\textbf{Network-file Queries}\newline TrafficNetQA \cite{Kwon2025TrafficNetQA:Files} & Transportation-network questions; repository-described resource & Tests network-structure interpretation & Full paper unavailable; evidence is limited to the inspected repository. \\
\addlinespace[3pt]
\textbf{Domain Knowledge}\newline TransportationGames \cite{Zhang2024TransportationGames:Models} & Textual/multi-modal tasks; context benchmark & Compares transportation-domain answers & External-database and real-time tasks are outside its evaluation. \\
\addlinespace[3pt]
\textbf{Itinerary Planning}\newline TripCraft \cite{Chaudhuri2025TripCraft:Planning} & Static itineraries; 1,000 queries in 140 cities (M2; transfer evidence) & Metric guidance raises the seven-day GPT-4o meal score from 0.61 to 0.80; final pass stays 0\% & Partial scores can hide infeasible plans; data access and redistribution are restricted. \\
\addlinespace[3pt]
\textbf{Route Planning}\newline MobilityBench \cite{Song2026MobilityBench:Scenarios} & 7,098 queries; deterministic API replay (M2) & ReAct/Gemini-3-Pro-Preview delivery: 84.38\%; valid final plans: 69.09\% & Replay excludes live API use; delivered answers can still fail constraints. \\
\bottomrule
\end{tabularx}
\end{table*}

\subsubsection{Data Integration \& Analytics}
\label{sub:integration}
LLMs extend transportation analytics to text-heavy sources alongside numerical inputs. Knowledge-management frameworks synthesize transportation information in a common analysis environment \cite{Somvanshi2024Gen-AIManagement}, while transit case studies show task-relevant retrieval from GTFS \cite{Jonnala2024UsingStudy}. Broader reviews identify opportunities to add textual context to forecasting and analytics \cite{Yan2025LargeOpportunities}. Ontology and knowledge-graph workflows organize entities and relations across these sources \cite{Tupayachi2024TowardsTransportation,Kuang2024HarnessingDecision-making}.

\rev{The comparative question is what the LLM component adds to the existing workflow. A heterogeneous system may expose only text or tool results to its LLM. Useful tests therefore separate retrieval, prediction and explanation, distinguish query execution from answer correctness, and preserve the availability of information at the intended decision time. This makes the incremental value of semantic context clearer without attributing a conventional predictor's performance to language reasoning.}

\subsubsection{Visualization \& Decision Support}
\label{sub:visualization}
One major application is scenario generation, where high-level natural-language descriptions are converted into simulation inputs \cite{Chang2024LLMScenario:Generation,Guzay2023AGeneration}. Detailed claims for the inaccessible G\"uzay report remain qualified in the evidence record. System-level orchestration extends this interface to running experiments and adjusting simulation settings \cite{Ye2025SUMO-MCP:Optimization}.

Language-guided simulation takes several forms. Collaborative agents support editable autonomous-driving scenes \cite{Wei2024EditableLLM-Agents}, while code generation can translate language instructions into guidance for a diffusion-based trajectory model \cite{Zhong2023Language-GuidedDiffusion}. Hierarchical reasoning supports controllable simulation \cite{Liu2024ControllableReasoning}, and textual context can assist policy transfer between simulators \cite{Da2024PromptLearning}. Personal mobility generation provides related simulation context \cite{Wang2024LargeGeneration}. \rev{These approaches differ in what the language model produces. Scene descriptions, code, parameters or actions, and visual plausibility should be complemented by task-specific operational checks.}

Tool orchestration offers a direct comparison with existing analytical practice. Reported traffic-simulation workflows reduce interaction time relative to command-line use, whereas smaller freight digital-twin examples mainly illustrate feasibility \cite{Ye2025SUMO-MCP:Optimization,Xu2025TowardsProtocol}. \phantomsection\label{par:calibration_evidence}\rev{Semantic parameter grouping also supports model calibration, although competitive rule-based grouping limits attribution of whole-framework improvements to the LLM alone \cite{Le2026LLMAssistedCalibration}. Reproducibility depends on retaining network and demand assumptions, tool settings and evaluation outputs, with agentic iteration bounded by the intended experiment and operator permissions.}

\subsubsection{Benchmarking \& Evaluation Platforms}
\label{sub:benchmarks}
Benchmarks assess consistency, robustness, constraint satisfaction and reproducibility alongside task accuracy under transportation-specific conditions.

Recent studies illustrate complementary evaluation targets. Transportation-domain questions assess knowledge and reasoning \cite{Syed2024BenchmarkingBehaviors,Zhang2024TransportationGames:Models}, while planning tasks combine domain knowledge with spatial and analytical requirements \cite{Ying2024BeyondPlanning}. Network-file question answering examines structured network understanding \cite{Kwon2025TrafficNetQA:Files}. Mobility-service benchmarks evaluate schedule-aware, preference-sensitive itineraries \cite{Chaudhuri2025TripCraft:Planning}, and route-planning benchmarks use real queries with deterministic tool replay \cite{Song2026MobilityBench:Scenarios}. Table~\ref{tab:eval} summarizes selected resources and their distinct evaluation purposes.

\rev{A common finding is the gap between component quality and complete task validity. Better partial itinerary scores can coexist with zero final pass rate, and a delivered response can still violate routing constraints. Knowledge-only questions, cached tool replay and closed-loop simulation therefore provide complementary evidence rather than interchangeable measures of readiness. Future TSMO benchmarks should combine these tests with clear task constraints, verified data access and separate evaluation of live operational effects.}

\section{Existing Gaps and Challenges}
\label{gaps}

\subsection{Data Heterogeneity and Structured Representation}
LLM, especially MM-LLM input integration must reconcile detector streams, connected-vehicle messages, incident logs, traveler feedback, images and video with different temporal/spatial resolution, semantics and reliability \cite{Shoaib2023ASystems}. Frequent numeric measurements may lack context, while rich textual and visual information must be aligned with current operational states.

\rev{Misaligned timestamps, sensor calibration errors and uncertain locations can place observations in the wrong traffic state. Calibration should be checked against reference measurements and coordinate mappings before fusion; residual error and observation age should accompany the state used for reasoning. Preserving source identities and conflicting observations supports operator verification.}

\subsection{Real-Time Inference and Operational Scalability}
Slow responses, unstable throughput and compute costs constrain agency deployment of LLMs, especially MM-LLMs. Shared resources across corridors or services can further delay time-sensitive incident response, traveler messages and signal decisions.

Real-time capability must therefore be defined for a specific decision. Static GTFS information queries \cite{Jonnala2025ExploringStudy} and live signal operations have different latency budgets. \rev{Acquisition, preprocessing, retrieval, tool execution, validation, and operator review can dominate the model's inference time. Table~\ref{tab:latency} separates the measured computational stages and their operational implications; averages alone conceal workload-dependent and tail delays.}
\begin{table*}[t]\reviewfloatcolor
\centering
\changedcolor
\caption{\rev{Selected computational timings and their measurement conditions.}}
\label{tab:latency}
\footnotesize
\setlength{\tabcolsep}{4pt}
\renewcommand{\arraystretch}{1.10}
\begin{tabularx}{\textwidth}{@{}>{\raggedright\arraybackslash}p{2.3cm} >{\raggedright\arraybackslash}X >{\raggedright\arraybackslash}X >{\raggedright\arraybackslash}X@{}}
\toprule
\textbf{Source} & \textbf{Reported Timing} & \textbf{Measurement Conditions} & \textbf{Operational Limit} \\
\midrule
LLMLight \cite{Lai2024LLMLight:Agents} & Inference: 3.53 s at batch 1; 5.31 s at batch 10 & Llama2-13B LightGPT; four A800 graphics processing units (GPUs), 80 gigabytes each & Excludes sensing, validation, communication and actuation. \\
\addlinespace[3pt]
VLMLight \cite{Wang2025VLMLight:Reasoning} & Reasoning pipeline: 9.35--11.48 s across three sites & Five A100 GPUs; simulated scenes & Compared with a 13 s phase window; deadline failures remain unmeasured. \\
\addlinespace[3pt]
Traffic-R1 \cite{Zou2025Traffic-R1:Systems} & Mean inference below 2 s & Centralized GPU server in the reported pilot & Edge execution and worst-case latency remain unmeasured. \\
\addlinespace[3pt]
IMPEL \cite{Nie2025JointApproach} & Forecast testing: 0.17 s; LLM encoding: 2.48 s at batch 1 & RTX A6000; separate prediction/encoding stages & Changing-input encoding cost and dispatch latency remain untested. \\
\bottomrule
\end{tabularx}
\end{table*}

\begin{table*}[t]\reviewfloatcolor
\centering
\changedcolor
\caption{\rev{Proposed task-dependent failure-to-assurance mapping; safeguards require validation for the intended action.}}
\label{tab:risk}
\footnotesize
\setlength{\tabcolsep}{4pt}
\renewcommand{\arraystretch}{1.10}
\begin{tabularx}{\textwidth}{@{}>{\raggedright\arraybackslash}p{2.8cm} >{\raggedright\arraybackslash}X >{\raggedright\arraybackslash}X >{\raggedright\arraybackslash}X@{}}
\toprule
\textbf{Failure Mode} & \textbf{Operational Consequence} & \textbf{Safeguard / Fallback} & \textbf{Evaluation Evidence} \\
\midrule
Wrong Event or Severity & Incorrect dispatch priority & Confirm critical fields; escalate uncertainty & Per-class recall; missed severe events; corrections \\
\addlinespace[3pt]
Wrong Location or Direction & Response reaches the wrong link & Check geographic information systems and roadway identifiers & Location/direction errors; ambiguous names \\
\addlinespace[3pt]
Unsupported Action & Infeasible route or unsafe phase & Independent constraints; approved controller fallback & Violations; unsafe proposals; rejection coverage \\
\addlinespace[3pt]
Fabricated Traveler Advice & Missed or inaccessible service & Source-linked templates; verify disruptions & Factual, accessibility and stale-data errors \\
\addlinespace[3pt]
Conflicting or Missing Inputs & Incorrect situation assessment & Preserve uncertainty; abstain or seek evidence & Sensor conflict; modality dropout; timestamp skew \\
\addlinespace[3pt]
Distribution Shift & Failure in unfamiliar conditions & Drift monitoring; calibrated rejection & Geographic/temporal holdouts; rare events; calibration \\
\addlinespace[3pt]
Deadline or Service Failure & Advice arrives too late & Timeout; legacy policy or operator escalation & Delay percentiles; uptime; recovery \\
\addlinespace[3pt]
Prompt Injection or Tool Misuse & Unauthorized access or action & Least privilege; sandbox; argument checks & Adversarial requests; privilege escalation; logs \\
\addlinespace[3pt]
Privacy Leakage & Exposed identities, trajectories or procedures & Access control; redaction; retention limits & Leakage tests; data lineage; permissions \\
\addlinespace[3pt]
Automation Bias & Unchallenged errors or unclear responsibility & Sign-off; training; audit and rollback roles & Error detection; workload; overrides; accountability \\
\bottomrule
\end{tabularx}
\end{table*}

\begin{table*}[t]\reviewfloatcolor
\centering
\changedcolor
\caption{\rev{Reported assurance mechanisms and additional operational verification needs.}}
\label{tab:assurance_evidence}
\footnotesize
\setlength{\tabcolsep}{4pt}
\renewcommand{\arraystretch}{1.10}
\begin{tabularx}{\textwidth}{@{}>{\raggedright\arraybackslash}p{2.3cm} >{\raggedright\arraybackslash}X >{\raggedright\arraybackslash}X >{\raggedright\arraybackslash}X@{}}
\toprule
\textbf{System / Source} & \textbf{Reported Mechanism} & \textbf{Evidence Supported} & \textbf{Verification Needed} \\
\midrule
iLLM-TSC \cite{Pang2024ILLM-TSC:Improvement} & Parse output; up to three attempts, then original RL action & Defined malformed-output fallback in simulation & Phase/timing feasibility; deadline failures \\
\addlinespace[3pt]
VLMLight \cite{Wang2025VLMLight:Reasoning} & Language-agent plan and legality review & Deliberative simulator checks & Independent rejection checks; workload stress tests \\
\addlinespace[3pt]
Traffic-R1 \cite{Zou2025Traffic-R1:Systems} & Expert plan approval; supervised incident adjustments & Reported oversight in a bounded live pilot & Independent outcomes; site failures; long-term operation \\
\addlinespace[3pt]
Road-situation/C-ITS \cite{Tong2025RoadSituationCITS} & Abstract Syntax Notation One (ASN.1) encoding and decoding checks & Verified message format; limited lane-status accuracy & Content accuracy; warning-delivery performance \\
\addlinespace[3pt]
TransitGPT \cite{Devunuri2024TransitGPT:Models} & Execution-error feedback; bounded retries & Better correctness on static-GTFS tasks & Code isolation; stale-data and service-failure tests \\
\addlinespace[3pt]
ARS \cite{Li2025ARS:Models} & Validate routes produced by generated programs & Validated-program success; optimality remains unresolved & Checker robustness; new constraints; live-dispatch effects \\
\addlinespace[3pt]
CTG++ \cite{Zhong2023Language-GuidedDiffusion} & Generated loss functions guide diffusion & Constraints influence generated autonomous-driving scenes & Code validation; physical/semantic scenario checks \\
\bottomrule
\end{tabularx}
\end{table*}

Hybrid architectures separate routine control from more expensive deliberation \cite{Wang2025VLMLight:Reasoning}. \rev{Their reliability depends on retaining controller deadlines and safety constraints across that interface. Model outages, retrieval failures, and timeouts create a further coordination problem: responsibility for continuing an approved policy, withholding an advisory, or escalating to an operator must remain clear.}

\subsection{Explainability, Trust, and Safety Assurance}
The black-box nature of MM-LLMs is a fundamental concern in safety-critical transportation. Operators are reluctant to rely on recommendations that cannot be explained, verified, or audited. Hallucinations, context drift, and inconsistency under unfamiliar conditions compound this problem \cite{Dona2024EvaluatingTasks}. Transportation AI assurance guidance identifies traceability, risk awareness, and operator oversight as core deployment requirements \cite{Yu2024UnderstandingProject}. Table~\ref{tab:risk} links operational failure modes to independent safeguards and evaluation evidence.

Table~\ref{tab:assurance_evidence} distinguishes reported mechanisms by the failure classes they address. \rev{iLLM-TSC's malformed-output fallback and VLMLight's model-based legality review provide different evidence from ARS's validation of generated routes. Output parsing, model self-review and independent feasibility checking should therefore be evaluated separately. Operational checks should enforce permitted signal phases and timing limits or route connectivity and fleet constraints before action authorization. }

\rev{Trust also depends on distinguishing retrieved facts, calculated outputs, and model-generated interpretation. Verbal confidence and plausible reasoning traces can encourage overreliance when factual support is weak. Confidence and attribution analyses in retrospective crash assessment \cite{Zhao2025SafeTrafficInterventions} address part of this problem; operational reliance additionally depends on whether operators detect and correct erroneous advice.}

\phantomsection\label{par:hallucination_comparison}\rev{Hallucination research distinguishes source-contradicting from unsupported generation and reviews lexical, entailment, question-answering, and human assessments \cite{Ji2023HallucinationSurvey}. These approaches reveal the limits of fluency and text overlap as reliability measures. In TSMO, errors in location, direction, event time, or priority can alter response selection, making action-relevant factual checks and missed-hazard assessment essential.}

\subsection{Multi-Modal Reasoning and Agentic System Integration}
Transportation operations require visual observations, textual reports, geospatial structure, and streaming sensor data to be interpreted consistently over time \cite{Le2024MultimodalSystems}. Integration becomes more demanding when an MM-LLM must coordinate tools, retrieve context, invoke models, and refine decisions. Agentic freight digital twins illustrate this emerging combination of language models, retrieval, and scientific tools \cite{Xu2025TowardsProtocol}. Its operational challenge is maintaining a coherent state across simulation, forecasting, control logic, and operator feedback.

Native vision-language interfaces \cite{Zhu2023MiniGPT-4:Models} and text-only agents using captions or serialized measurements expose different information to reasoning. \rev{Unclear interfaces can propagate lost context or misunderstood outputs across agents. Multi-agent failure research identifies coordination and verification problems \cite{Cemri2025WhyFail}, while transportation assurance guidance highlights common-mode failures in redundant components \cite{Yu2024UnderstandingProject}. Adding agents therefore increases the importance of explicit task boundaries and independently checked handoffs.}

\leavevmode\phantomsection\label{par:toolformer_comparison}\rev{Toolformer learns API selection and arguments by retaining calls that improve subsequent token prediction, evaluated through zero-shot language tasks and perplexity \cite{Schick2023Toolformer}. Transportation orchestration instead depends on consistency across multi-step, state-changing workflows. Toolformer's reported limits in chained or interactive use and tool-dependent call cost highlight unresolved requirements for argument validity, failed-call recovery, authorization, and deadline compliance.}

\subsection{Governance, Standardization, and Institutional Barriers}
Data privacy concerns can make agencies hesitant to share incident logs, traveler data, or operational transcripts with centralized AI services \cite{Yao2024AUgly}. Incompatible schemas, APIs, and evaluation workflows also hinder transfer across agencies. Institutional adoption depends on combining operational utility with assurance, oversight, and deployment context \cite{Yu2024UnderstandingProject}; fragmented responsibility and validation practices increase integration and maintenance costs.

\rev{Model-level privacy evaluation, informed by the LLM-PBE privacy benchmark \cite{Li2024LLM-PBE:Models}, must be connected to the handling of prompts, tool outputs, embeddings, and audit logs. Local hosting leaves retention, access, redaction, and cross-agency sharing obligations in place. Accountability requires documented responsibilities for data stewardship, model/tool maintenance, action approval, incident review and rollback. For liability assessment, records should connect model and tool versions, recommendations and approval decisions so that provider, agency and operator contributions can be examined. Allocation of legal liability requires separate assessment under the applicable institutional and legal arrangements.}

\phantomsection\label{par:crossfire_comparison}\rev{MalwareBench in LLMs Caught in the Crossfire evaluates malicious code requests using refusal and graded response-content measures, comparing automated judges with expert labels \cite{Li2025MalwareBench}. For transportation assistants, this distinction extends to direct unauthorized requests and instructions embedded in retrieved incident reports or webpages. Treating such content as operator authority can expose tools to unintended actions. Cybersecurity evaluation should therefore test whether injected instructions can trigger unauthorized tool calls or expose protected data, and whether least-privilege access, argument checks and code isolation block these paths and support safe recovery.}

\phantomsection\label{par:puo_comparison}\rev{The Panel Understanding and Operation benchmark (PUO-Bench) evaluates appliance-panel description, grounding, functions, and planning; its privacy framework keeps image parsing local and sends descriptions to cloud reasoning \cite{Lin2025PUOBench}. Its static-image planning setting highlights two transfer issues for traffic cameras and operator consoles: residual identity or location disclosure in transmitted text, and grounding as operational states change. Data minimization must be assessed together with task completion and timing.}

\begin{figure*}[!t]\reviewfloatcolor
\centering
	\includegraphics[width=\textwidth]{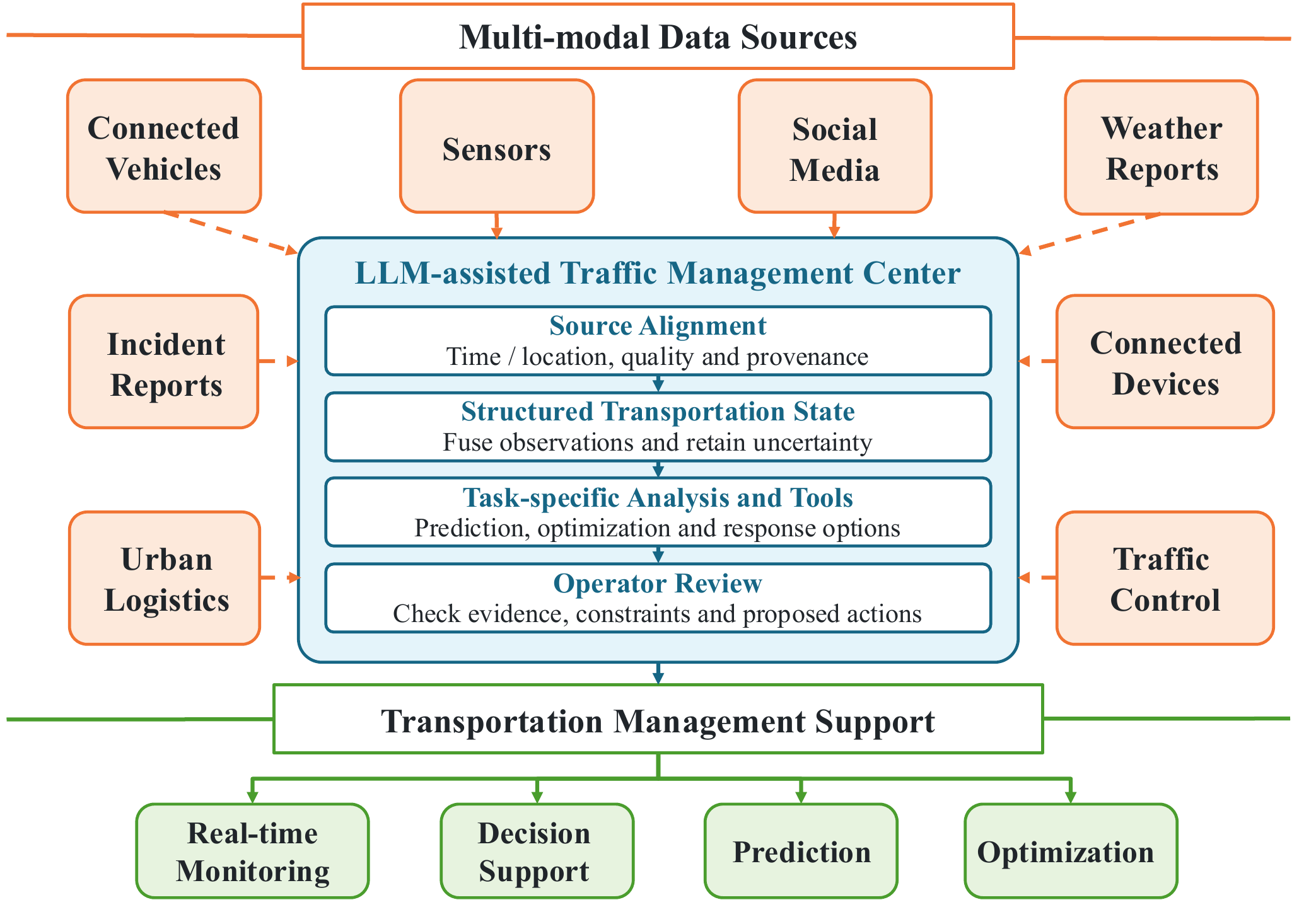}
\caption{\rev{Proposed traffic management center workflow. Source alignment and a structured transportation state support task-specific analysis, response options and operator review before information reaches operational users.}}
\label{Fig:TMC_Elements}
\end{figure*}

\begin{figure*}[!t]\reviewfloatcolor
\centering
	\includegraphics[width=\textwidth]{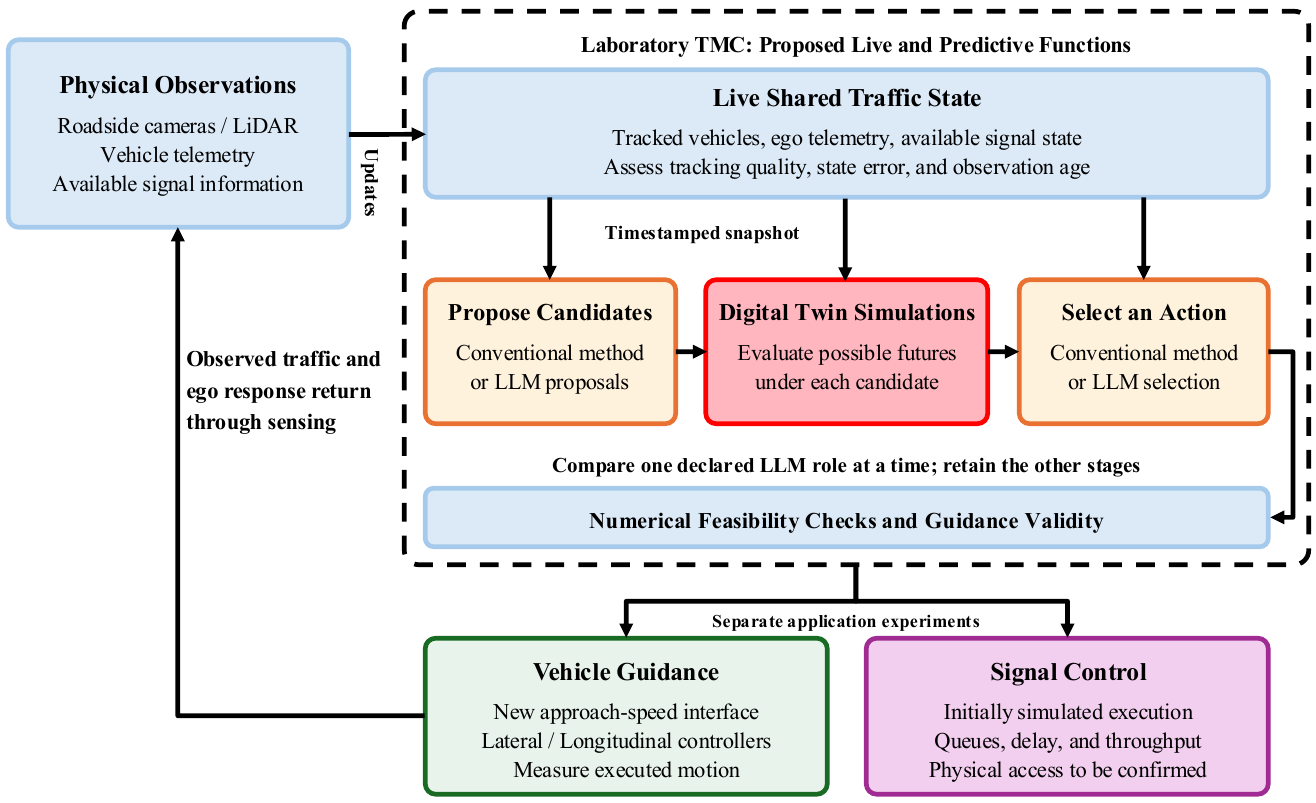}
\caption{\rev{Proposed LLM-assisted digital-twin TMC. Blue: conventional/shared functions; yellow: LLM-enabled decision stages; red: candidate-action simulations initialized from the shared traffic state; green: ego-vehicle guidance; purple: infrastructure control. }}
\label{Fig:proposed_system}
\end{figure*}

\section{Future Directions}
\label{future}
Research should establish where LLM and MM-LLM capabilities improve existing workflows and under which operating conditions. \rev{Grounded information services offer bounded near-term tasks, while recommendations and increasingly automated control require progressively stronger tests of feasibility, timing, and recovery.} \rev{Figures~\ref{Fig:TMC_Elements} and~\ref{Fig:proposed_system} present proposed designs that provide a basis for future implementation and evaluation.}

\subsection{Structured Multi-modal Data Alignment for Transportation Context}
General multi-agent workflow and infrastructure research provides supporting context for structured multi-modal alignment \cite{Li2024AChallenges}. Figure~\ref{Fig:TMC_Elements} shows how shared context representations could connect observations with transportation-management functions.

One promising direction is to combine MM-LLMs with knowledge graphs, event graphs, and domain-specific ontologies, grounding unstructured observations in machine-interpretable entities and relations \cite{Kuang2024HarnessingDecision-making}. Transportation-aware context engineering can build on related smart-city agent work \cite{Kalyuzhnaya2025LLMSystems}. Where training data remain distributed across agencies, federated fine-tuning offers a complementary research direction \cite{Yan2025FederatedDirections}, with privacy, communication cost, and prediction performance evaluated together.

\rev{Matched numerical-only, text-only, visual-only, and combined-input tests should measure each modality's incremental value under geographic and temporal holdouts. Missing cameras, stale reports, and contradictory observations provide useful stress cases.} Evaluation should identify how native modality encoders and text-conversion interfaces affect information loss, ambiguity, and state consistency.

\subsection{Real-Time Inference and Operational Scalability}
Hierarchical deployment can place lightweight models or routing modules at the edge and selectively invoke more capable reasoning for complex decisions. Quantization, pruning, distillation, and hardware-aware optimization \cite{Paul2025EdgeReview} should be studied jointly with workload partitioning, event-triggered invocation, and operator escalation.

For intersection-management applications \cite{Masri2024LeveragingScenarios}, research should compare selective reasoning with continuous invocation under matched workloads. \rev{The stage-specific measurements in Table~\ref{tab:latency} motivate end-to-end and tail-latency testing with declared hardware, concurrency, input size, and network conditions.} \rev{Outage and timeout experiments should assess whether approved fallback policies maintain continuity while the existing controller's deadlines and safeguards remain authoritative.} Operational value, robustness, and compute cost should guide the placement of reasoning within the architecture.

\subsection{Trustworthy, Explainable, and Human-Centered LLM Decision Support}
Retrospective traffic-safety work on domain adaptation and attribution provides a starting point \cite{Zhao2025SafeTrafficInterventions}. \rev{The next priority is to test explanation faithfulness, confidence calibration, and appropriate operator reliance in the tasks where advice will be used.} Fairness, auditability, and life-cycle monitoring should be incorporated into these evaluations \cite{Wandelt2024LargeChallenges,Yu2024UnderstandingProject}.

\rev{Bounded pilots should compare source-linked summaries, information extraction, or constrained queries with an existing interface on the same tasks. Relevant outcomes include unsupported statements, error detection, correction time, and operator workload.} Ambiguous place names, stale records, and out-of-scope questions can test escalation behavior. Recommendation studies should also measure feasibility and adherence to agency procedures, with approval responsibilities defined in advance.

\subsection{Agentic Multi-modal Integration Frameworks for TSMO}
Emerging transportation and logistics architectures provide examples of planning and digital-twin interaction \cite{Xu2025TowardsProtocol}. General agentic multi-modal surveys \cite{Yao2025AModels} and cooperative autonomous-driving research \cite{You2024V2X-VLM:Models} offer architectural guidance that requires adaptation to agency workflows.

\rev{Figure~\ref{Fig:proposed_system} proposes a digital-twin-based (DT) separate vehicle-guidance and signal-control experiments using shared, timestamped traffic states and candidate simulations. Matched inputs and simulation budgets isolate LLM proposal or selection. Infrastructure-assisted ego guidance retains the lateral and longitudinal controllers; signal execution is initially simulated. Feasibility checks precede action; integration remains proposed. Tables~\ref{tab:traffic} and~\ref{tab:public} provide task-specific gains and trade-offs; Table~\ref{tab:assurance_evidence} identifies additional assurance needs.} 

\rev{Implementation should isolate generated code, restrict tool permissions, and require independent constraint checks and authorization before consequential actions. Versioned logs should link inputs, tool calls, validation outcomes, and approved actions.} \rev{A staged evaluation program can progress through controlled replay, stress-tested simulation, shadow operation without actuator access, and bounded supervised use. Held-out tests and versioned review should precede feedback-driven model updates.}

\subsection{Benchmarking, Standardization, and Cross-Agency Validation}
Task-level sensor reconstruction at different compression ratios provides one evaluation example \cite{Yang2024TransCompressor:Transportation}. Future benchmarks should connect such component measures with robustness to distribution shift, multi-modal consistency, explanation quality, and operator acceptability. Common incident schemas and modular APIs can reduce integration costs and support cross-agency testing.

The resources in Table~\ref{tab:eval} provide complementary evaluation tasks \cite{Syed2024BenchmarkingBehaviors,Kwon2025TrafficNetQA:Files,Chaudhuri2025TripCraft:Planning,Song2026MobilityBench:Scenarios}. Cross-agency comparisons should distinguish offline replay, simulated closed-loop control, and live operation. \rev{Shared records should specify permitted actions, data and tool versions, matched baselines, geographic and temporal holdouts, operator-study design, and recovery outcomes. Reporting repeated-run variation and disclosing retrieval, tool-access, and compute budgets would improve comparability.} Linking these tests to procurement and governance processes can support responsible transfer across agencies.

\section{Conclusions}
\label{conclusions}
This survey synthesizes LLM applications in TSMO, linking operational tasks to model roles, validation evidence and practical requirements. The literature supports an assistive role in integrating heterogeneous data, summarizing information and supporting natural-language operator interaction. \rev{Across prototypes, simulation and author-reported live settings, usefulness depends on latency, multi-modal alignment, explainability, governance and benchmarking.}

Source-specific comparisons reveal gains and trade-offs: improved task accuracy can increase response time, while reduced simulated waiting time may coexist with higher travel time. Cross-agency transfer requires attention to supervision, operating conditions and comparison design. \rev{The proposed LLM-assisted DT framework connects shared traffic states, candidate simulations and checked guidance; future evaluation should isolate the LLM contribution under matched conditions.}

\section*{Acknowledgments}
This study is partially funded by the “Medium SoCal OASIS Internal Funding Award” at University of California, Riverside and the National Center for Sustainable Transportation (NCST) UTC Program.

\bibliographystyle{IEEEtran}
\bibliography{references_edison,reference_style}

\clearpage

\section{Biography Section}

\vspace{-33pt}

\begin{IEEEbiography}[{\includegraphics[width=1in,height=1.25in,clip,keepaspectratio]{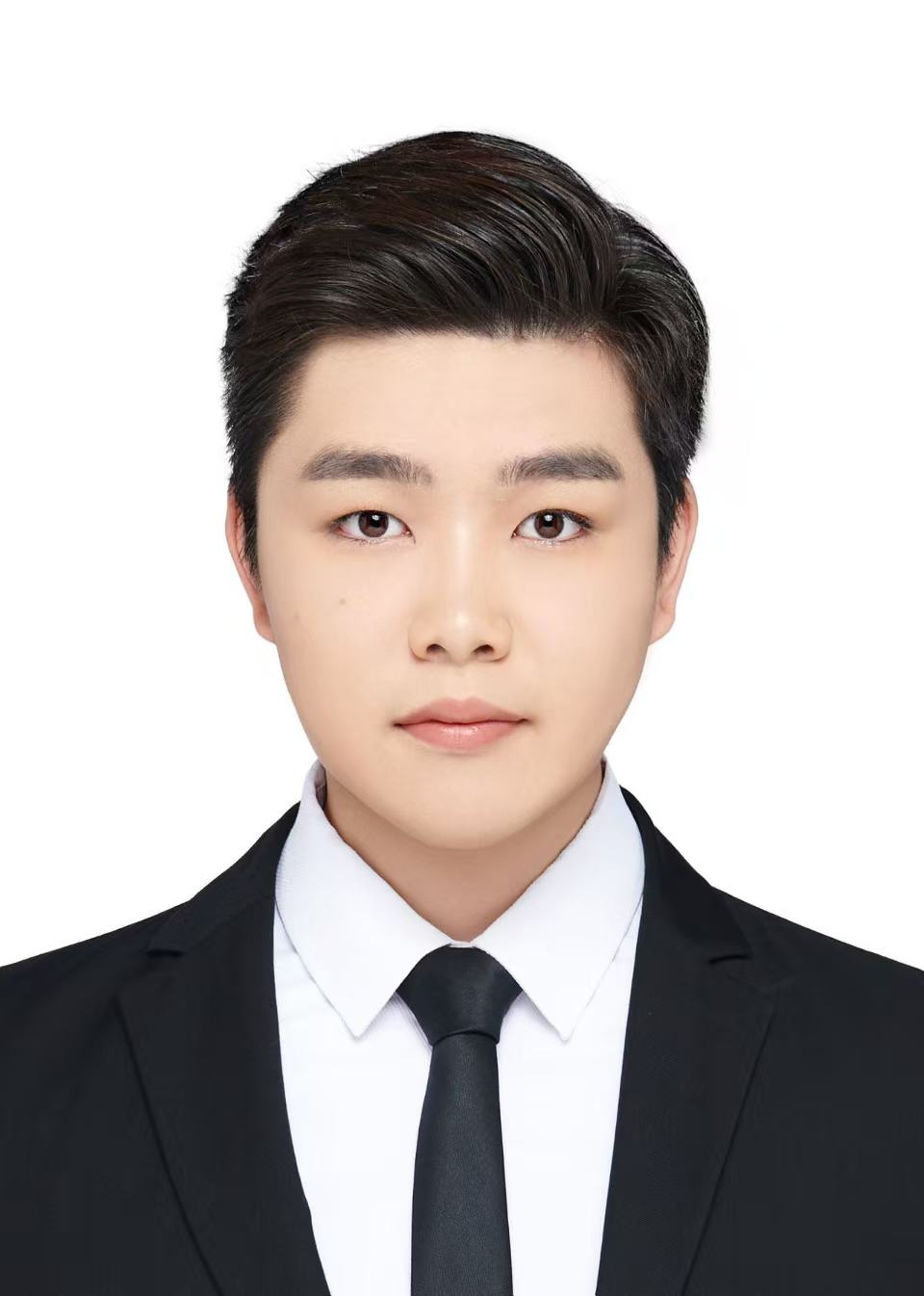}}]{Siyan Li}
(Student Member, IEEE) is a Ph.D. student at the Bourns College of Engineering--Center for Environmental Research and Technology (CE-CERT), University of California, Riverside, Riverside, CA, USA. He received the bachelor's degree in Electrical Engineering and Automation from China Agricultural University, Beijing, China, and the master's degree in Electrical and Computer Engineering from the University of California, Riverside, Riverside, CA, USA. His research focuses on large language models for transportation, prediction, and connected and automated vehicles, developing AI-enabled methods that integrate foundation models, multi-modal sensing, and simulation for intelligent mobility.
\end{IEEEbiography}

\begin{IEEEbiography}[{\includegraphics[width=1in,height=1.25in,clip,keepaspectratio]{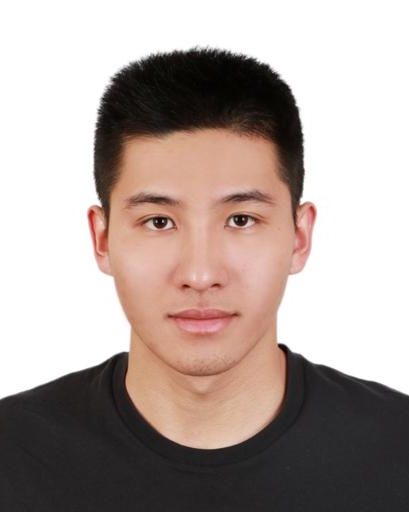}}]{Zehao Wang}
(Student Member, IEEE) is a Ph.D. student in Computer Science at the University of California, Riverside, Riverside, CA, USA. He received the B.S. degree in Software Engineering from Sun Yat-sen University, Guangzhou, China, in 2020, and the M.S. degree in Computer Science from New York University, New York, NY, USA, in 2023. His research interests include autonomous navigation, multi-agent systems, vision-language models, vision-language-action models, human-robot interaction, and autonomous driving. His current work focuses on developing communication-efficient and robust multi-agent algorithms for collaborative perception, prediction, and decision-making in real-world autonomous systems.
\end{IEEEbiography}

\begin{IEEEbiography}[{\includegraphics[width=1in,height=1.25in,clip,keepaspectratio]{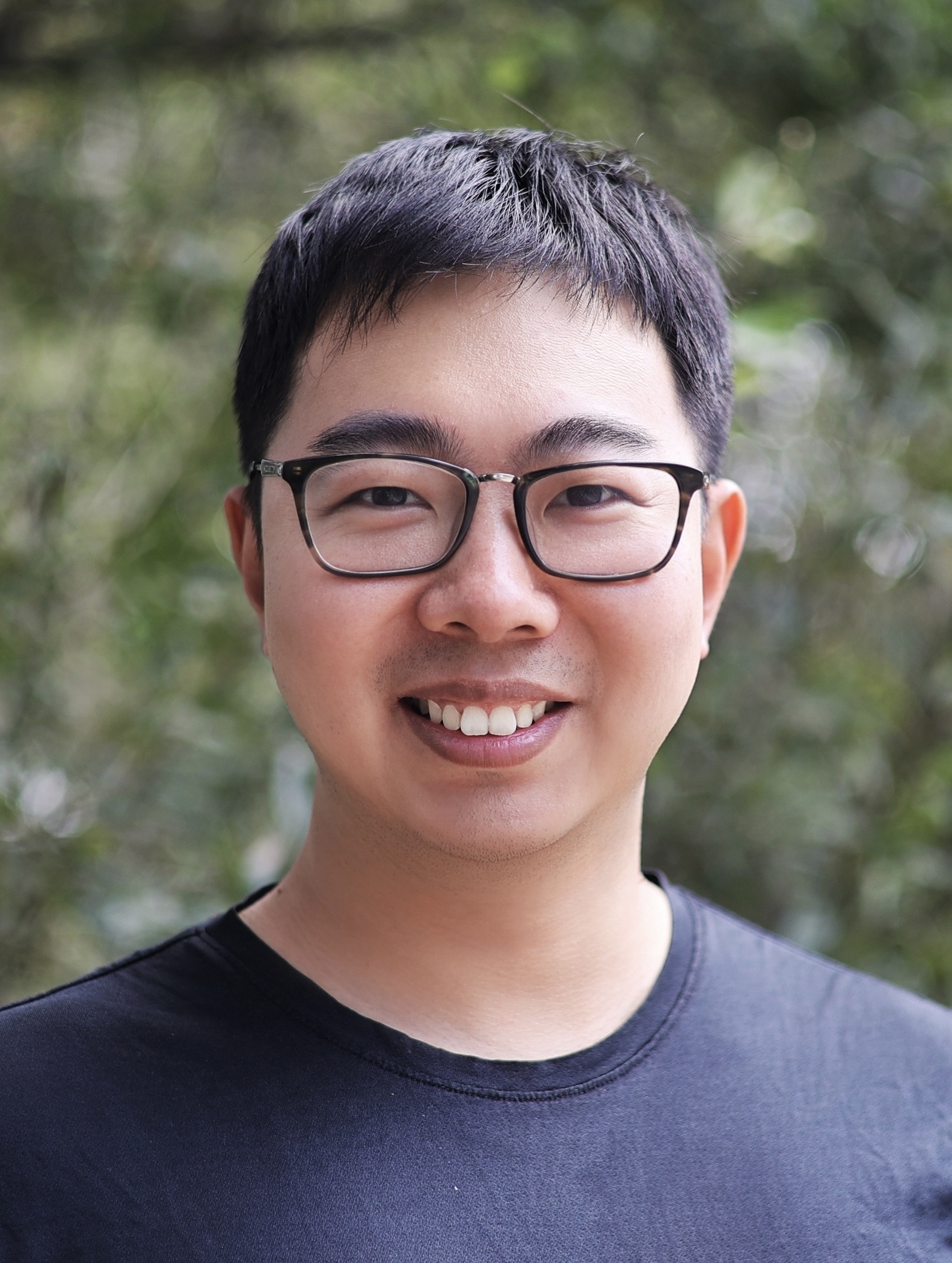}}]{Jiachen Li}
(Member, IEEE) is an Assistant Professor in the Department of Electrical and Computer Engineering and, by courtesy, the Department of Computer Science and Engineering at the University of California, Riverside. He leads the Trustworthy Autonomous Systems Laboratory and is a core faculty member of the Riverside Artificial Intelligence Research Institute and the Center for Robotics and Intelligent Systems. Before joining the University of California, Riverside, he was a Postdoctoral Scholar at Stanford University and received his Ph.D. from the University of California, Berkeley. Dr. Li was named a Robotics: Science and Systems (RSS) Robotics Pioneer and an American Society of Mechanical Engineers Dynamic Systems and Control Division Rising Star. He currently serves as Co-Chair of the IEEE Robotics and Automation Society Technical Committee on Robot Learning and as an Associate Editor for IEEE Transactions on Robotics, IEEE Robotics and Automation Letters, and an Area Chair for multiple leading conferences. He has also organized multiple workshops at top international venues. His research interests span robotics, trustworthy artificial intelligence and machine learning, foundation models, reinforcement learning, control, optimization, and computer vision, with a focus on intelligent autonomous systems operating in human-centered and multi-agent environments.
\end{IEEEbiography}

\vfill
\newpage

\begin{IEEEbiography}[{\includegraphics[width=1in,height=1.25in,clip,keepaspectratio]{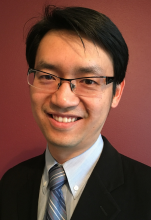}}]{Kanok Boriboonsomsin}
(Member, IEEE) received his Ph.D. degree in transportation engineering from the University of Mississippi, Oxford, MS, USA, in 2001. He is a Full Research Engineer in the College of Engineering - Center for Environmental Research and Technology (CE-CERT) and a Full Adjunct Professor in the Department of Electrical and Computer Engineering at the University of California, Riverside, CA, USA. His research interests include sustainable transportation, transportation energy and emission modeling, eco-friendly intelligent transportation systems, and connected and automated vehicles, with a focus on freight transportation. Dr. Boriboonsomsin currently serves as an Associate Editor for IEEE Intelligent Transportation Systems Magazine, and as a member of the Transportation Research Board's Transportation Energy Data and Technology.
\end{IEEEbiography}

\begin{IEEEbiography}[{\includegraphics[width=1in,height=1.25in,clip,keepaspectratio]{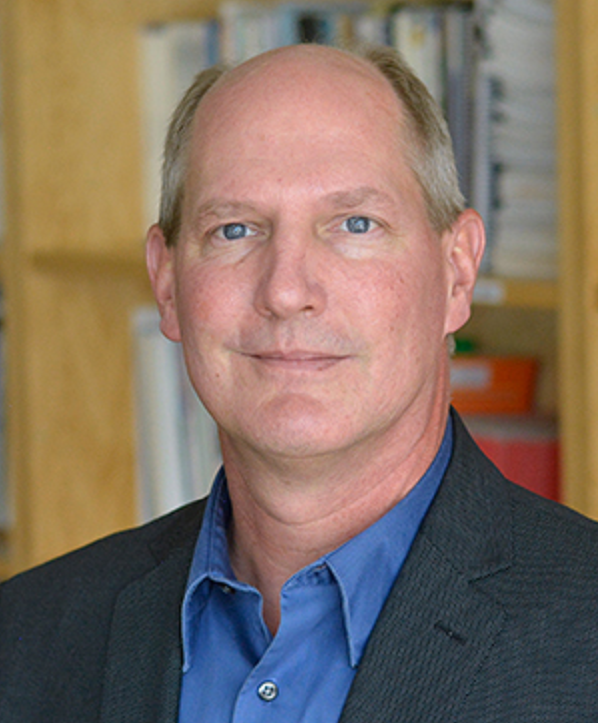}}]{Matthew J. Barth}
(Fellow, IEEE) received his M.S. and Ph.D. degrees in Electrical and Computer Engineering from the University of California at Santa Barbara in 1985 and 1990, respectively. He is currently the Esther \& Daniel Hays Distinguished Professor and Associate Dean of Research and Graduate Education at the Bourns College of Engineering, University of California-Riverside. He is part of the intelligent systems faculty in Electrical and Computer Engineering and conducts his research at the Center for Environmental Research and Technology (CE-CERT), the University of California, Riverside’s largest multi-disciplinary research center. Dr. Barth’s research focuses on Intelligent Transportation Systems, with the goal of improving environmental sustainability. His current research interests and teaching portfolio includes sustainable transportation, connected and automated vehicles, cooperative perception systems, advanced navigation, shared mobility, and vehicle electrification. Barth has been very active in the IEEE Intelligent Transportation Systems Society (ITSS) for many years, serving as a past Senior Editor for both IEEE Transaction on Intelligent Transportation Systems and the IEEE Transaction on Intelligent Vehicles. Barth participated in a variety of roles in the IEEE ITSS, including Society President from 2014-2015. He has received the IEEE ITSS Outstanding Research Award in 2017 and the IEEE ITS Institution Lead Award in 2020. He is an IEEE Fellow, and a Fellow of the National Academy of Inventors.
\end{IEEEbiography}

\begin{IEEEbiography}[{\includegraphics[width=1in,height=1.25in,clip,keepaspectratio]{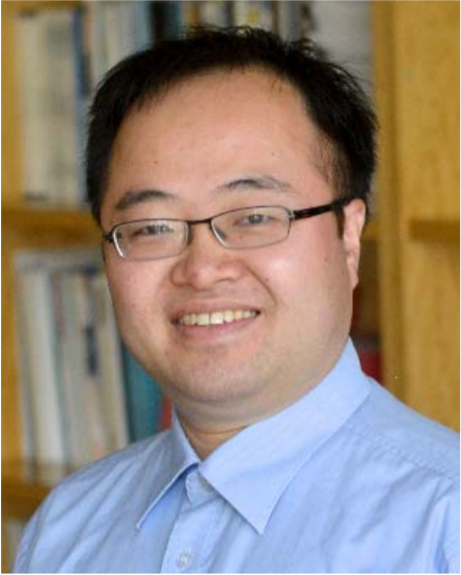}}]{Guoyuan Wu}
(Senior Member, IEEE) received his Ph.D. degree in mechanical engineering from the University of California, Berkeley in 2010. Currently, he holds an Associate Researcher and an Associate Adjunct Professor position at Bourns College of Engineering – Center for Environmental Research \& Technology (CE–CERT) and Department of Electrical \& Computer Engineering in the University of California at Riverside. His research lies in development and evaluation of sustainable and intelligent transportation system technologies, including connected and automated transportation systems, shared mobility, transportation electrification, optimization and control of vehicles, traffic simulation, and emissions measurement and modeling. Dr. Wu serves as Associate Editors for a few journals, including IEEE Transactions on Intelligent Transportation Systems, SAE International Journal of Connected and Automated Vehicles, and IEEE Open Journal of ITS. He is also a member of the Vehicle-Highway Automation Standing Committee (ACP30) of the Transportation Research Board, a board member of Chinese Institute of Engineers Southern California Chapter, and a member of Chinese Overseas Transportation Association. He is a recipient of Vincent Bendix Automotive Electronics Engineering Award.
\end{IEEEbiography}

\vspace{11pt}

\vfill

\end{document}